\documentclass[lettersize,journal]{IEEEtran}
\usepackage{amsmath,amsfonts,amssymb}
\usepackage{algorithmic}
\usepackage{algorithm}
\usepackage{array}
\usepackage[caption=false,font=normalsize,labelfont=sf,textfont=sf]{subfig}
\usepackage{textcomp, gensymb}
\usepackage{stfloats}
\usepackage{url}
\usepackage{verbatim}
\usepackage{graphicx}
\usepackage{cite}
\usepackage{comment}
\usepackage{color}
\usepackage{colortbl}
\usepackage{float}
\usepackage{multirow}
\usepackage{wrapfig}
\usepackage{overpic}
\usepackage{booktabs}
\usepackage{ragged2e}

\usepackage[pagebackref,breaklinks,colorlinks]{hyperref}

\usepackage[switch]{lineno}

\usepackage{subcaption}

\usepackage{amsmath,amsfonts,bm}

\newcommand{\tabref}[1]{Table~\ref{#1}}

\def\figref#1{figure~\ref{#1}}
\def\Figref#1{Figure~\ref{#1}}

\def\Secref#1{Section~\ref{#1}}

\def\eqref#1{equation~\ref{#1}}

\def\1{\bm{1}}

\DeclareMathAlphabet{\mathsfit}{\encodingdefault}{\sfdefault}{m}{sl}
\SetMathAlphabet{\mathsfit}{bold}{\encodingdefault}{\sfdefault}{bx}{n}

\newcommand{\myPara}[1]{\vspace{.05in}\noindent\textbf{#1.}}

\graphicspath{{./Imgs/}}

\newcommand{\moduleNameFull}{\textbf{L}ightweight \textbf{G}aussian \textbf{A}sset \textbf{A}dapter}
\newcommand{\moduleName}{LGAA}

\usepackage{xspace}
\makeatletter
\DeclareRobustCommand\onedot{\futurelet\@let@token\@onedot}
\def\@onedot{\ifx\@let@token.\else.\null\fi\xspace}

\def\etc{\emph{etc}\onedot}

\makeatother

\ifCLASSINFOpdf
\else
\fi

\definecolor{myRed}{rgb}{0.8,0.2,0.2}
\definecolor{myBlue}{rgb}{0.2,0.2,0.8}
\definecolor{myBlack}{rgb}{0.0,0.0,0.0}

\begin{document}

\title{
\textbf{DreamLifting}: A Plug-in Module Lifting MV Diffusion Models for 3D Asset Generation
}

\author{Ze-Xin~Yin, Jiaxiong~Qiu, Liu~Liu, Xinjie~Wang, Wei~Sui, Zhizhong~Su, Jian~Yang, and~Jin~Xie
\thanks{Z.X. Yin (zexin.yin.cn@mail.nankai.edu.cn) and J. Yang are with PCA Lab, VCIP, College of Computer Science, Nankai University.}
\thanks{J.Xie is with School of Intelligence Science and Technology, Nanjing University, Suzhou, China.}
\thanks{J. Qiu, L. Liu, X. Wang, Z. Su are with Horizon Robotics.}
\thanks{W. Sui is with D-Robotics.}
\thanks{This work is partially done while Z.X. Yin is an intern at Horizon Robotics \& D-Robotics.}
\thanks{J. Xie is the corresponding author (csjxie@nju.edu.cn).}}

\markboth{Journal of \LaTeX\ Class Files,~Vol.~xxx, No.~xxx, xxx~xxx
\quad\quad\quad\quad
\quad\quad\quad\quad
\quad\quad\quad\quad
\quad\quad\quad\quad
\quad\quad\quad\quad
DOI:~\url{https://doi.org/10.1109/TVCG.2026.3679384}\quad}%
{Shell \MakeLowercase{\textit{et al.}}: A Sample Article Using IEEEtran.cls for IEEE Journals}

\IEEEpubid{0000--0000/00\$00.00~\copyright~xxx IEEE}

\maketitle

\begin{figure*}[ht]    
    \subfloat[Text- and image-conditioned generation results.]{
        \begin{overpic}[width=0.485\linewidth]{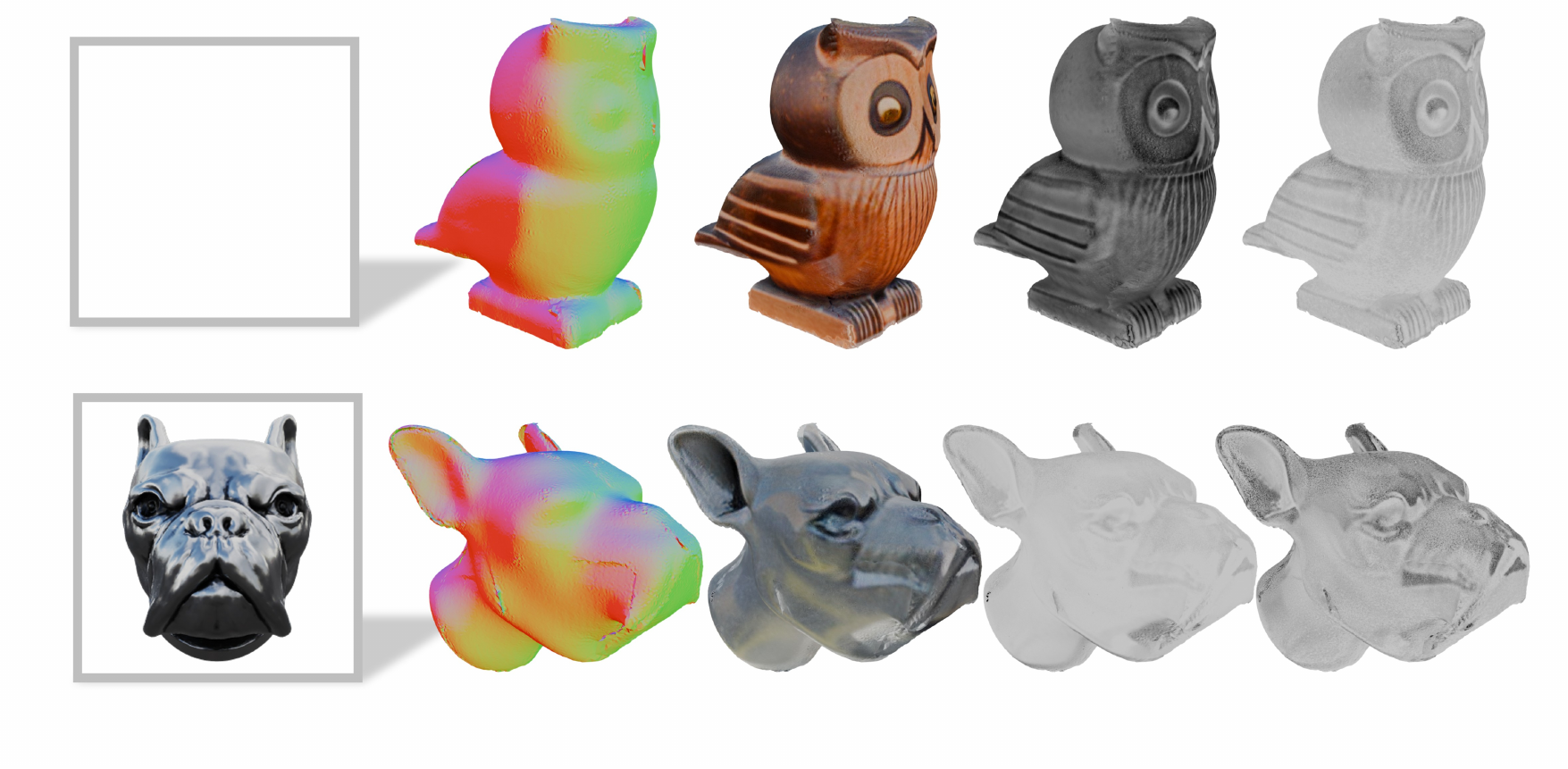}
            \put(6, 42){\textit{A wooden }}
            \put(6, 37){\textit{owl, 3d }}
            \put(6, 32){\textit{model.}}
            \put(9, 1){\small input}
            \put(28, 1){\small normal}
            \put(47, 1){\small albedo}
            \put(64.2, 1){\small metallic}
            \put(80.5, 1){\small roughness}
        \end{overpic}
    }
    \hfill
    \subfloat[Generated 3D assets under different HDRI maps.]{
        \includegraphics[width=0.485\linewidth]{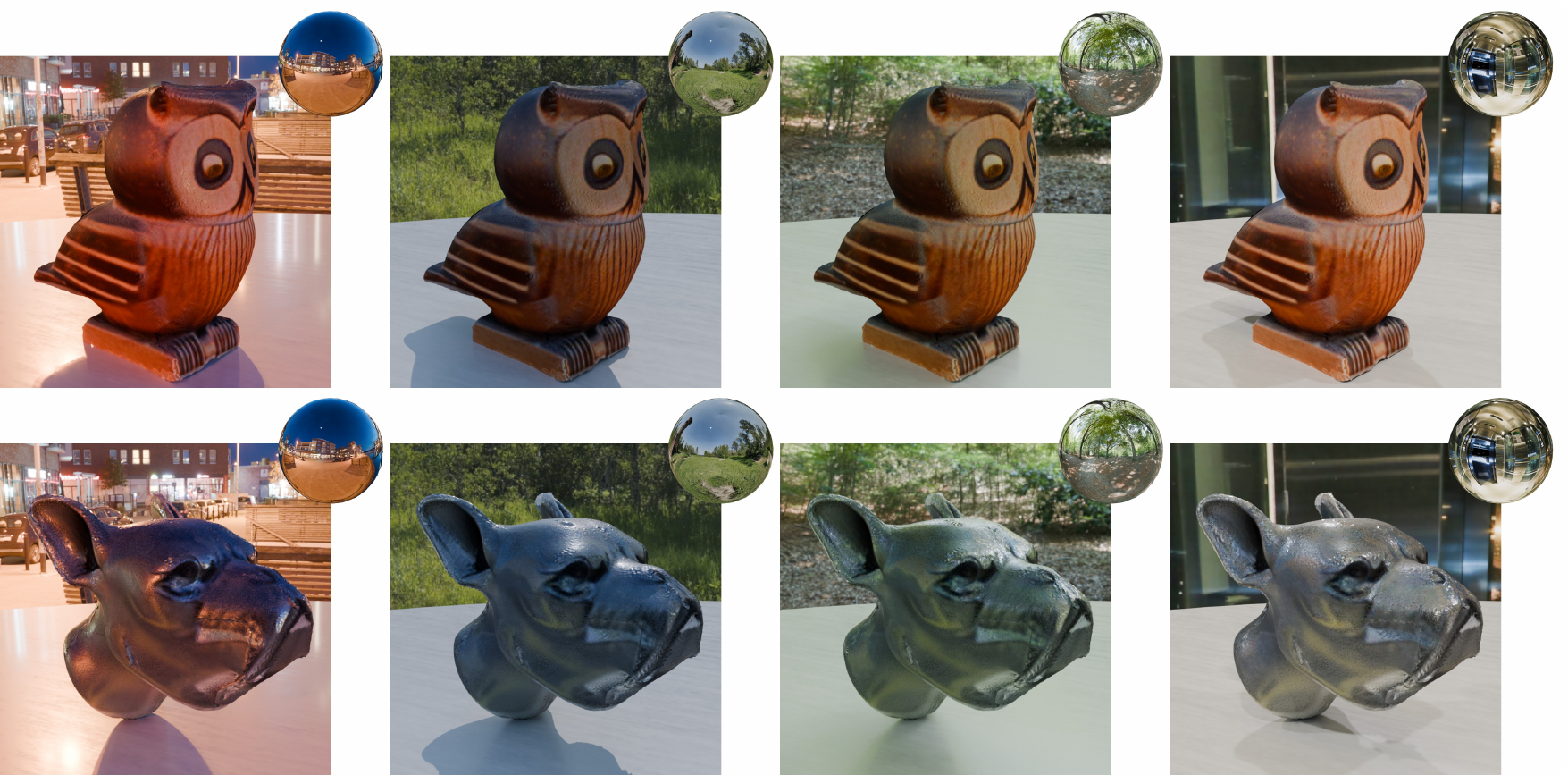}
    }
    \vfill
    \subfloat[Multiple 3D assets generated by our pipeline in different environment maps.]{
        \includegraphics[width=0.485\linewidth]{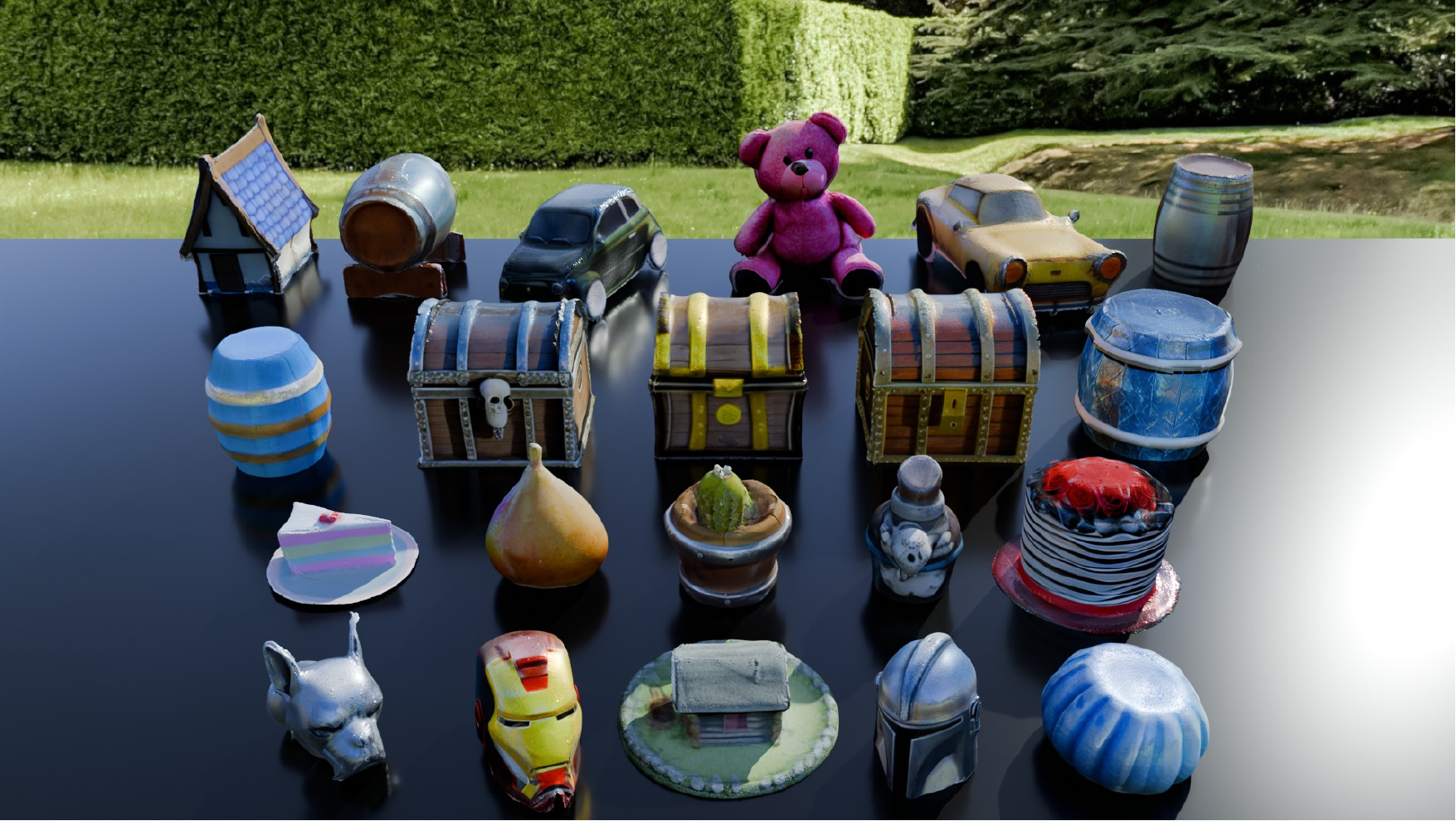}
        \hfill
        \includegraphics[width=0.485\linewidth]{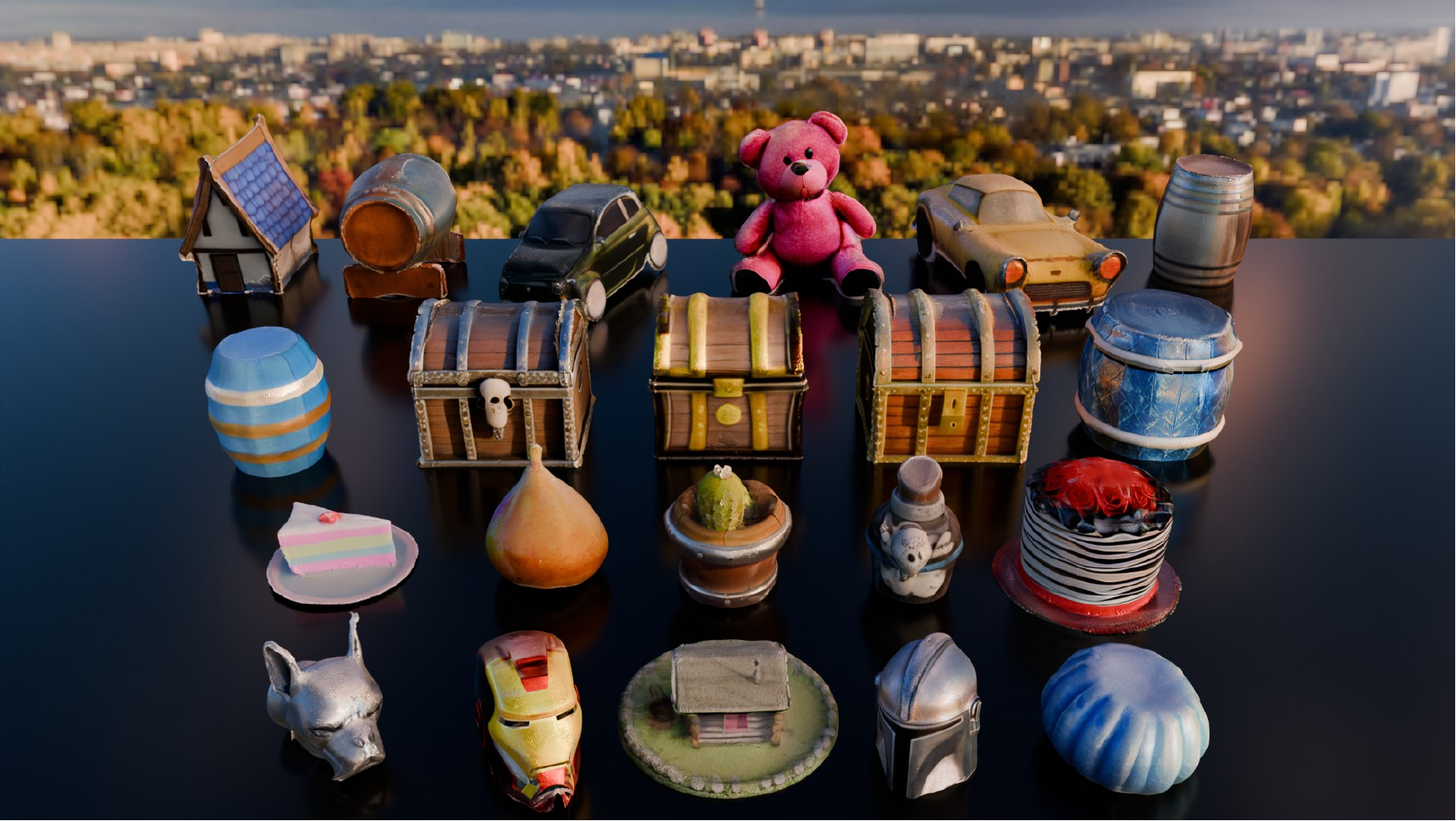}
    }
    \caption{
        Our pipeline possesses the capability of generating diverse, PBR-ready 3D assets from either text prompts or image conditions.
        The synthesized assets are fully relightable with accurate PBR materials; for example, the wooden owl instance exhibits 
        diffuse color changes under different environment maps, while the specular dog instance successfully reflects its surroundings. 
        These results highlight the usability of the generated 3D assets.
    }
    \label{fig:teaser}
\end{figure*}


\begin{abstract}
    The labor- and experience-intensive creation of 3D assets with physically based rendering (PBR) materials demands an autonomous 3D asset creation pipeline.    
    However, most existing 3D generation methods focus on geometry modeling, either baking textures into simple vertex colors or leaving texture synthesis
    to post-processing with image diffusion models.
    To achieve end-to-end PBR-ready 3D asset generation, we present \moduleNameFull~(\moduleName), a novel framework that unifies the modeling of geometry and PBR materials
    by exploiting multi-view (MV) diffusion priors from a novel perspective.
    The \moduleName~features a modular design with three components. 
    Specifically, the \moduleName~Wrapper reuses and adapts network layers from MV diffusion models, which encapsulate knowledge acquired from billions of images,
    enabling better convergence in a data-efficient manner.
    To incorporate multiple diffusion priors for geometry and PBR synthesis, the \moduleName~Switcher aligns 
    multiple \moduleName~Wrapper layers encapsulating different knowledge.
    Then, a tamed variational autoencoder (VAE), termed \moduleName~Decoder, is designed to predict 2D Gaussian Splatting (2DGS) with PBR channels.
    Finally, we introduce a dedicated post-processing procedure to effectively extract high-quality, relightable mesh assets from the resulting 2DGS.
    Extensive quantitative and qualitative experiments demonstrate the superior performance of \moduleName~with
    both text- and image-conditioned MV diffusion models.
    Additionally, the modular design enables flexible incorporation of multiple diffusion priors,
    and the knowledge-preserving scheme effectively preseves the 2D priors learned on massive image dataset,
    which leads to data efficient finetuning to lift the MV diffuison models for 3D generation with merely 69k multi-view instances.
    Our code, pre-trained weights, and the dataset used will be publicly available via our project page:
        \url{https://zx-yin.github.io/dreamlifting/}.
\end{abstract}




\section{Introduction}
\label{sec:intro}

\IEEEPARstart{M}{odern} graphics pipelines rely on high-quality, PBR-ready 3D assets to achieve photorealistic rendering, 
particularly in film, game production, autonomous driving, virtual reality, \etc. 
However, the creation of such assets by human artists demands significant labor and expertise. 
This raises a strong demand for AI-driven content generation methods in the field of 3D asset creation.

Although recent advances in native 3D generation~\cite{yang2024hunyuan3d, zhang2024clay, zhao2025hunyuan3d,
yang2025holopart,he2025sparseflex,li2025triposg, wu2025direct3ds2gigascale3dgeneration, li2025sparc3d}
have produced highly detailed geometries, these methods still fail to model vivid PBR textures.
Early works, including distillation-based approaches~\cite{poole2022dreamfusion,wang2023score, chen2023fantasia3d, 
lin2023magic3d, wang2024prolificdreamer, chen2024text, tang2023dreamgaussian} and multi-view generation–reconstruction pipelines~\cite{shi2023zero123++, long2024wonder3d, liu2023syncdreamer, 
shi2023mvdream,li2023instant3d, hong2023lrm, xu2023dmv3d, xu2024instantmesh, tang2024lgm, xu2024grm}, typically bake textures as simple vertex colors, 
which do not support relighting or photorealistic rendering.
While these methods can leverage multi-view material diffusion models to synthesize PBR textures 
and project them onto UV maps, such operations introduce additional issues such as misalignment or blurring in the texture maps. 
3DTopia-XL~\cite{chen20243dtopia} compresses SDF and PBR materials into a specially designed 3D representation named PrimX 
and builds a latent diffusion model upon PrimX, resulting in an end-to-end 3D asset generation framework.
However, it is challenging to satisfy the need for large quantities of high-quality, PBR-ready 3D assets required to train 3DTopia-XL.
Therefore, the community requires new data-efficient paradigms for PBR-ready 3D asset generation.

To explore a new paradigm for 3D asset generation, we introduce a novel perspective on utilizing 2D MV diffusion models.
Previous works~\cite{huang2024mv,long2024wonder3d,shi2023zero123++} demonstrate that MV RGB diffusion models 
can synthesize multi-view consistent images, from which 3D meshes can be extracted using neural 3D reconstruction methods~\cite{wang2021neus,mildenhall2021nerf},
indicating the implicit geometry priors encapsulated in the MV RGB diffusion models.
Furthermore, subsequent studies~\cite{qiu2024richdreamer,li2024idarb,liang2025diffusionrenderer,sun2025depth} extend diffusion models to 
generate multi-view consistent surface normal maps, depth maps, and PBR materials, 
which inherently encode valuable PBR priors. 
These pioneering works underscore the encapsulated geometric and PBR priors in MV diffusion models, making them well suited for high-quality 3D asset creation.

\IEEEpubidadjcol

\begin{figure}[ht]    
    \begin{overpic}[width=\linewidth]{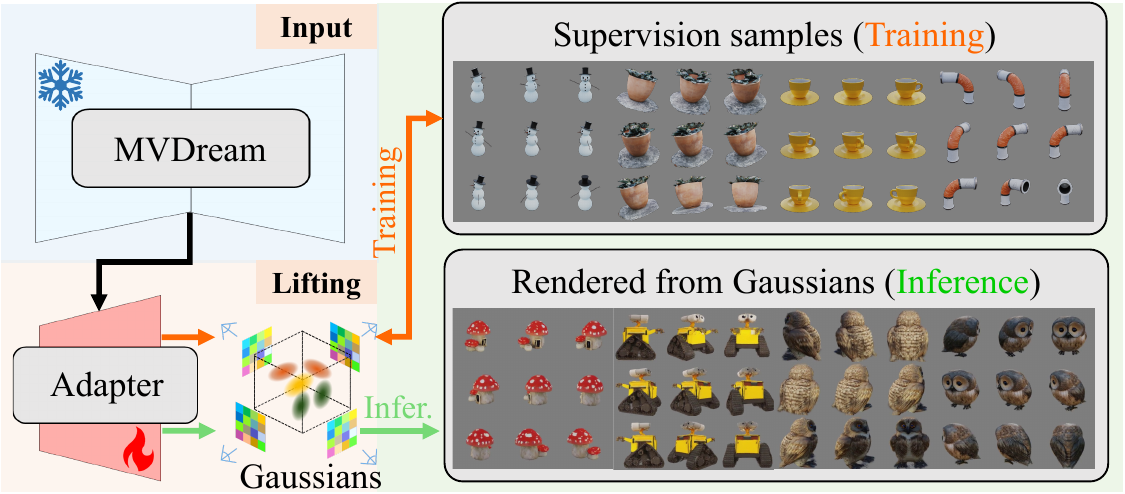}
    \end{overpic}
    \caption{
       A toy experiment showing that intermediate features from the pretrained MV diffusion are 
       geometry-aware and can be lifted to an explicit 3D representation with minimal adaptation.
       We construct the adapter by reusing the upsampling blocks from the Unet of MVDream v2.1, 
       modify the output head to predict pixel-aligned 3DGS, and finetune the adapter on a subset of G-Objaverse~\cite{deitke2023objaverse}.
       This indicates that MV diffusion models contain strong geometry cues that can be lifted to an explicit 3D representation with minimal adaptation.
       Details about the toy experiment can be found in the Supplementary.
    }
    \label{fig:insight}
\end{figure}

However, existing MV diffusion–based 3D generation pipelines still rely on large reconstruction models (LRMs) or neural reconstruction methods to generate 3D assets.
Neural reconstruction methods suffer from time-consuming per-prompt optimization,
whereas LRMs enable feed-forward 3D asset generation from multi-view images but require large amounts of high-quality 3D data for training.
Beyond these two paradigms, we are interested in \textit{whether pre-trained MV diffusion models already encode geometry priors that can be 
directly converted into an explicit 3D representation.}
In \Figref{fig:insight}, we conduct a toy experiment showing that the intermediate feature maps from a pre-trained MV diffusion model
can be directly converted into 3DGS via an adapter, which is modified from the upsampling blocks of the MV diffusion Unet.
The results provide evidence that pre-trained MV diffusion models encode strong 3D priors beyond multi-view consistent image synthesis, 
and these priors can be efficiently converted to 3D assets via carefully designed lifting modules.

In this paper,
we introduce the \moduleNameFull~(\moduleName) to explore the geometric and PBR priors
encoded in MV diffusion models for end-to-end, high-quality, PBR-ready 3D asset generation from a novel perspective.
The \moduleName~features a plug-in module that integrates seamlessly with pre-trained multi-view diffusion.
Specifically, we first design a simple yet effective \moduleName~Wrapper module, which serves as the core component 
for adaptively preserving and fusing pre-trained knowledge for 3D asset generation.
The network layers of MV diffusion models encode robust knowledge acquired from billions of images,
and the \moduleName~Wrapper clones and freezes these pre-trained layers, and injects learnable, 
zero-initialized convolutional layers to adapt this knowledge for 3D asset generation.
To incorporate multiple priors, including MV RGB diffusion priors and MV PBR material diffusion priors,
for geometry and PBR material generation,
we introduce the \moduleName~Switcher, which aligns different priors in a layer-wise manner using learnable, zero-initialized convolutional layers,
which prevents conflicts between priors during the early training stage and enables the progressive and adaptive growth of alignment.
Finally, we propose the \moduleName~Decoder to generate pixel-aligned 2D Gaussian Splatting (2DGS)~\cite{kerbl3Dgaussians,huang20242d, szymanowicz2024splatter}
as the 3D representation that bridges 2D priors and 3D content creation.
During training, as our module predicts PBR-ready 2DGS in a feed-forward manner, 
supervision can be performed solely using rendered RGB images and corresponding G-buffer images.
The 2D supervision scheme facilitates the incorporation of various image-based loss functions.
Therefore, we introduce an image-based differentiable deferred shading scheme to closely align the
generated PBR materials with their corresponding image appearances during training, substantially enhancing their realism.

The design of \moduleName~offers several advantages: 
(1) It adaptively preserves expressive knowledge learned from large-scale 2D datasets and utilizes it for 3D generation,
enabling efficient and high-quality 3D asset synthesis.
Our experiments show that preserving and adapting these pre-trained priors facilitates better convergence with less training data.
(2) It maintains modularity and flexibility, allowing seamless integration with different base models and enabling scalable performance improvements 
when paired with more powerful foundational diffusion models.
Extensive experiments confirm that integrating more powerful base models enhances generation quality 
and that flexibly combining PBR diffusion priors significantly improves material synthesis.

For downstream applications, we implement a test-time refinement process to convert the resulting 
Gaussian splats into high-quality, UV-mapped 3D meshes, as shown in \Figref{fig:teaser}.
At inference, the system generates Gaussian splats with PBR materials in under three seconds, 
which are then efficiently converted into production-ready assets by the proposed process within 30 seconds.

Our contributions are as follows: 
\begin{itemize}
\item We introduce Lightweight Gaussian Asset Adapter (\moduleName), a modular and plug-in approach that leverages MV diffusion models 
for direct Gaussian asset synthesis without extensive, large-scale 3D data.
\item We present a simple yet effective scheme to exploit 2D diffusion priors for 3D asset generation, 
offering several advantages, including improved performance without relying on large-scale 3D data 
and scalable generation quality when integrated with more powerful backbone models.
\item We propose a feed-forward framework with post-processing, enabling the rapid generation of high-quality, 
PBR-ready 3D assets within 30 seconds.
\end{itemize}


\section{Related works}
\label{sec:related_works}

\begin{figure*}[ht]
\centering
    \begin{overpic}[width=0.95\linewidth]{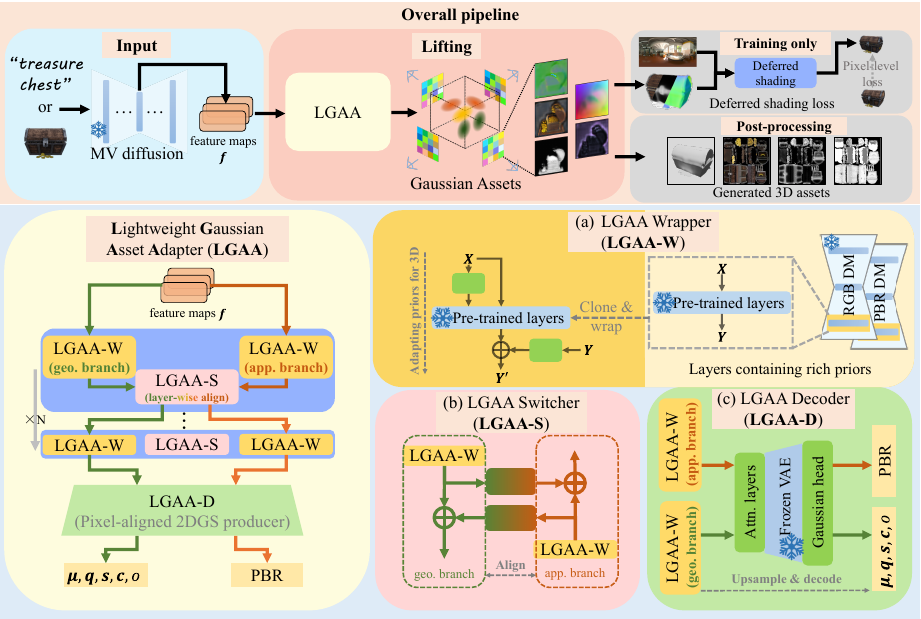}    
    \small

    \put(49.8, 36){$\mathcal{ZC}$}
    \put(58.1, 28.6){$\mathcal{ZC}$}

    \put(54, 14.2){$\mathcal{ZC}$}
    \put(54, 10.3){$\mathcal{ZC}$}
    
    \end{overpic}
    \caption{
    Overall of the 3D asset generation pipeline. We propose the \moduleNameFull~(\moduleName), which is composed of three components:
    (a) \moduleName~Wrapper (\moduleName-W), (b) \moduleName~Switcher (\moduleName-S), and (c) \moduleName~Decoder (\moduleName-D),
    where $\mathcal{ZC}$ indicates zero-initialized convolutional layers. 
    In (a), we adapt the priors for 3D generation by wrapping pre-trained layers with $\mathcal{ZC}$ layers, where $\bm{X}$ indicates input feature maps, 
    $\bm{Y}$ are feature maps from MV diffusion models, and $\bm{Y}'$ are output maps from the \moduleName-W. 
    We wrap layers from RGB DM (diffusion model) to construct the geometry branch, and use those from PBR DM for the appearance branch.
    In our implementations, we use MVDream/ImageDream/DreamView as the RGB DMs, and use IDArb as the PBR DM.
    In (b), we align the geometry and appearance branches with $\mathcal{ZC}$ layers, which progressively develop bidirectional information exchange paths during training.
    In (c), the \moduleName-D upsample the feature maps, decodes PBR channels of albedo $\bm{a} \in \mathbb{R}^3$,
    metallic $m \in \mathbb{R}$, and roughness $r \in \mathbb{R}$ for the appearance branch,
    and decodes Gaussian parameters of 3D position $\bm{\mu} \in \mathbb{R}^3$, rotation quaternion $\bm{q} \in \mathbb{R}^4$,
    scaling vector $\bm{s} \in \mathbb{R}^2$, opacity $o \in \mathbb{R}$ and color $\bm{c} \in \mathbb{R}^3$ for the geometry branch.
    During the training procedure, we tie the G-buffer maps with the RGB images via image-based deferred shading.
    In inference, we extract the 3D mesh with PBR material maps from the Gaussian Splat assets with carefully designed post-processing.}
    \label{fig:pipe} 
\end{figure*}

\myPara{Optimization-based 3D generation with diffusion priors}
With the development of diffusion theory~\cite{ho2020denoising} and the emergence of 
various neural 3D representations~\cite{mildenhall2021nerf, wang2021neus, shen2023flexicubes, kerbl3Dgaussians, shen2021deep, mescheder2019occupancy},
3D generation has achieved significant progress in terms of quality and speed.
Dreamfusion~\cite{poole2022dreamfusion} proposes Score Distillation Sampling (SDS) loss 
to distill 3D consistent NeRF from text-to-image (T2I) diffusion priors given only text prompts,
which opens up a new era for zero-shot 3D generation and 
follow-up efforts that improve distillation 
quality~\cite{yu2023text, liang2024luciddreamer, wang2024prolificdreamer}, 
apply on different neural representations~\cite{chen2023fantasia3d, chen2024text, lin2023magic3d, yi2023gaussiandreamer, tang2023dreamgaussian},
perform scene-level generation~\cite{zhang2024gsctrl,fang2023ctrl},
and even extend to 4D generations~\cite{ren2023dreamgaussian4d, singer2023text, ling2024align},
occur like mushrooms after rain.
However, due to the lack of 3D priors in T2I diffusion models, 
distillation-based methods suffer from the 3D inconsistency problem, also known as the Janus problem;
therefore, the community paves the way to inject multi-view 3D priors into T2I models 
by fine-tuning pre-trained models using rendered multiple views from large-scale dataset~\cite{deitke2023objaverse, deitke2024objaverse}
to generate multi-view consistency images~\cite{shi2023mvdream, liu2023syncdreamer, long2024wonder3d, shi2023zero123++, zhang2025ar, qiu2024richdreamer, li2023sweetdreamer,yan2024dreamview},
which server as strong 2D and 3D combined priors and improve the generation quality of distillation-based methods by a large margin.
However, this diagram suffers from time-consuming per-prompt optimization.

\myPara{Feed-forward 3D generation with diffusion priors}
To solve the limitations of distillation-based methods,
instant3D~\cite{li2023instant3d} proposes a new diagram to decompose the text-to-3D generation task 
into text-to-MV images generation and MV-to-3D generation tasks,
the former phase is implemented as a fine-tuned multi-view 2D diffusion model, 
while the latter phase features a feed-forward network mapping multi-view images to NeRF representation.
The two-stage diagram indirectly benefits from both 2D and 3D priors
and demonstrates superiority against previous methods regarding quality and speed.
InstantMesh~\cite{xu2024instantmesh} and CRM~\cite{wang2024crm} extend the diagram to direct mesh generation, 
while LGM~\cite{tang2024lgm}, GRM~\cite{xu2024grm}, LaRa~\cite{chen2024lara}, and Turbo3D~\cite{huturbo3d} build the reconstruction models with Gaussian Splats.
As the models are typically large transformer-like networks with millions of learnable parameters,
the training process either consumes tens to hundreds of high-end GPUs or requires a large-scale dataset to converge,
which is typically unaffordable in the academic community.
Besides, these methods all model the appearance of the generated 3D instances as simple vertex colors,
failing to synthesize PBR materials; therefore, the applications of generated 3D models are limited.

Recently, diffusion-based native 3D generation methods have achieved promising results.
DiffSplat~\cite{lin2025diffsplat} explores the reuse of
pre-trained text-to-image diffusion models 
to produce Gaussian splat images, achieving 
good rendering quality, but they do not decouple 
PBR materials and geometry.
TRELLIS~\cite{xiang2025structured} is able to model complex topology via proposed structured latents,
but the representation tends to model oversaturated vertex colors.
3DShape2VecSet~\cite{10.1145/3592442} paves the way to generate high-quality shapes via a 3D diffusion model,
and following methods~\cite{li2024craftsman, zhao2025hunyuan3d, wu2025direct3ds2gigascale3dgeneration, li2025sparc3d,chen2025dora}
are able to generate much more detailed geometry.
But these methods all focus on geometry modeling without PBR materials.

\myPara{3D generation with PBR materials}
Simultaneously recovering geometry, materials, and illumination is a highly ill-posed problem 
even with densely captured data~\cite{liu2023nero, zhu2024gs, zhang2021nerfactor,li2024tensosdf,zhu_2025_dsdf};
therefore, most existing 3D generation works synthesize meshes with simple vertex colors or even without textures,
which are incompatible with the modern graphics pipeline.
Several works~\cite{liu2023unidream, xu2023matlaber, qiu2024richdreamer} introduce PBR priors into 
the optimization process to achieve material decomposition during generation,
but these methods suffer from the time-consuming per-prompt optimization.
Meta 3D AssetGen~\cite{siddiqui2024meta} proposes a feed-forward network to regress SDF fields 
with coarse PBR materials and refine the textures using a texture refiner network,
while ARM~\cite{feng2024arm} introduces a feed-forward network to generate a differentiable mesh representation
and leverages a triplane-based PBR synthesizer to regress the PBR materials.
However, these two-stage methods increase the complexity of the generation pipeline.
TexGaussian~\cite{xiong2025texgaussian} proposes a pipeline to generate Gaussian assets with PBR channels 
with 3D meshes as the conditions,
while TexGen~\cite{yu2024texgen} generates PBR materials for 3D meshes in the UV space.
However, these methods rely on high-quality 3D meshes as input.
3DTopia-XL~\cite{chen20243dtopia} introduces a novel 3D representation with the ability to encode PBR material channels, enabling text- / image-to-3D asset generation.
However, it requires large amounts of computing resources and high-quality, PBR-ready 3D data for training.

Additionally, several recently released datasets are facilitating the development of 3D asset generation.
Objects-with-lighting~\cite{Ummenhofer2024OWL} introduces a real-world dataset containing object-level multi-view images captured under different environment maps;
however, this dataset contains only eight objects.
DTC~\cite{dong2025digital} presents a large-scale digital twin dataset comprising 2,000 real-world objects, 
each with detailed PBR materials, geometry, and associated multi-view DSLR photographs captured under diverse environment maps.
MAGE~\cite{wang2025mage} filters the Objaverse dataset, leading to a subset containing 17k 3D instances,
and renders corresponding G-buffers and images from multiple views for each filtered instance.
IDArb~\cite{li2024idarb} releases a large-scale dataset containing 5.7 million multi-view posed RGB images with PBR materials, 
significantly promoting the development of PBR asset generation.



\section{Methods}
\label{sec:methods}

An overview of our framework is illustrated in \Figref{fig:pipe},
which is designed for 3D asset generation via extracting geometric and appearance priors from 2D diffusion models
without requiring access to large amounts of high-quality 3D data.
First, we briefly review the 3D Gaussian Splat representation and the multi-view diffusion models that constitute our backbone in \Secref{subsec:bkgd}.
Next, we detail the architecture of \moduleName~and describe its integration with various multi-view diffusion models in \Secref{subsec:module}.
To facilitate 3D asset generation with PBR materials, 
we introduce an image-based differentiable deferred rendering scheme that 
links intrinsic components to the final rendered appearance in \Secref{subsec:pbr_rend}.
Finally, we outline a detailed procedure for extracting 3D meshes with PBR materials from Gaussian assets in \Secref{subsec:post}.

\subsection{Preliminaries: Gaussian Splats and Multi-view diffusion models}
\label{subsec:bkgd}

\myPara{Pixel-aligned Gaussian Splats}
3D Gaussian Splatting (3DGS)~\cite{kerbl3Dgaussians} proposes  
to parameterize the 3D scene via radiance fields 
in the form of a collection of Gaussian primitives $\mathcal{G}=\{\bm{g}_i\}$,
where each primitive contains multiple attributes recovered by differentiable rendering.
2D Gaussian Splatting (2DGS)~\cite{huang20242d} further improves the 3DGS for accurate geometry representation,
each primitive of which is parameterized by 
a 3D position $\bm{\mu} \in \mathbb{R}^3$, a rotation quaternion $\bm{q} \in \mathbb{R}^4$,
a scaling vector $\bm{s} \in \mathbb{R}^2$, an opacity $o \in \mathbb{R}$, 
and a view-dependent appearance $\bm{SH} \in \mathbb{R}^{(d+1)^2 \times 3}$ 
represented by spherical harmonic of degree $d$.
As an explicit representation, 2DGS is an unstructured representation,
which is incompatible with traditional 2D neural networks.
Inspired by Splatter Image~\cite{szymanowicz2024splatter},
we leverage 2DGS and organize them in the form of multi-view splatter images,
where each pixel represents one Gaussian primitive;
therefore, we can easily leverage 2D neural networks to generate 2DGS 
with accurate geometry.
\myPara{Multi-view diffusion model}
Multi-view (MV) diffusion models are typically fine-tuned from stable diffusion~\cite{rombach2022high} 
to generate 3D consistent MV images.
MVDream~\cite{shi2023mvdream} and DreamView~\cite{yan2024dreamview} are fine-tuned to 
generate four orthogonal views around the object
with elevation at the range of $[0\degree, 30\degree]$ through the denoising process.
ImageDream~\cite{wang2023imagedream} finetunes MVDream with carefully designed pixel injection scheme
to achieve image-to-multi-view generation.
IDArb~\cite{li2024idarb} is a multi-view-RGB-conditioned diffusion model 
to decouple multi-view consistent, physically based rendering materials.
The above diffusion models contain rich 2D and 3D priors,
based on which we conduct our experiments.

\subsection{\moduleNameFull}
\label{subsec:module}

Current MV diffusion models are capable of generating multi-view consistent RGB appearance and PBR materials,
demonstrating their encapsulation of geometric and PBR material priors.
To leverage these rich priors for direct 3D asset generation, we propose our \moduleNameFull,
which adaptively preserves priors from MV diffusion models, adapts them for 3D asset generation,
and generates Gaussian splats with PBR attribute channels.
As shown in \Figref{fig:pipe}, \moduleName~comprises three fundamental components:
\moduleName~Wrapper, \moduleName~Switcher, and ~\moduleName~Decoder.
Detailed descriptions of each component are provided below.

\myPara{\moduleName~Wrapper}
Current MV diffusion models are trained on billions of high-quality images, and thus their network layers encapsulate robust knowledge.
Our \moduleName~Wrapper is designed to exploit the encapsulated knowledge for our task.
Therefore, the module must preserve the pre-trained layers 
while also being able to adapt their outputs.
We achieve this in a simple yet effective manner.

Following \cite{zhang2023adding}, we define a \textit{network block} in diffusion models as modular units 
commonly utilized in neural network architectures, including ResNet blocks, convolution batch-normalization ReLU (conv-bn-relu) blocks, 
transformer blocks, among others.
As illustrated in \Figref{fig:pipe}(a), the \moduleName~Wrapper consists of layers cloned and frozen from the middle and upsampling 
blocks of the UNet architecture.
These layers typically receive an input feature map $\bm{X}_{in}$, a skip connection feature $\bm{X}_{res}$ (absent in the middle layer), 
and additional conditional information $\bm{C}$, to produce an output feature map $\bm{Y}$.
Given such a network block $\mathcal{F}(\bm{X}_{in}, \bm{X}_{res}, \bm{C}; \Theta)$ with pre-trained parameters $\Theta$, 
our \moduleName~Wrapper creates a parallel copy of this block, adapts the information flow with zero-initialized $1\times1$ convolutional layer $\mathcal{ZC}(\cdot)$,
and operates as follows:
\begin{equation}
    {
    \begin{array}{lcl}
    \bm{Y} & = & \mathcal{F}(\bm{X}_{in}, \bm{X}_{res}, \bm{C}; \Theta) \\
    \bm{Y}' & = & \mathcal{F}(\bm{X}_{in}, \bm{X}_{res} + \mathcal{ZC}(\bm{X}_{res}), \bm{C}; \Theta) + \mathcal{ZC}(\bm{Y})
    \end{array}}
\end{equation}
This design maximizes the preservation of pre-trained priors while enabling adaptability for 3D asset generation.

\myPara{\moduleName~Switcher}
\moduleName~Wrapper offers the flexibility to integrate multiple priors,
for example, we can create parallel branches, each encapsulating knowledge from different MV diffusion models
to separately synthesize geometry (mean positions $\bm{\mu}$ of Gaussian primitives) and appearance (other Gaussian Splat attributes).
However, this creates the need to align information flow from the parallel branches to ensure consistent instance generation.
To enhance consistency and alignment between geometry and appearance features, 
we propose a lightweight \moduleName~Switcher, implemented using zero-initialized $1\times1$ convolutional (ZC) layers, as illustrated in \Figref{fig:pipe}(b).
Specifically, given a geometry branch layer $\mathcal{F}_g(\cdot)$ and an appearance branch layer $\mathcal{F}_a(\cdot)$,
which produce feature maps $\bm{Y}_g$ and $\bm{Y}_a$, respectively,
the \moduleName~Switcher enables bidirectional information exchange between these branches as follows:
\begin{equation}
    \begin{array}{ccc}
    \bm{Y}'_g & = & \bm{Y}_g + \mathcal{ZC}(\bm{Y}_a) \\
    \bm{Y}'_a & = & \bm{Y}_a + \mathcal{ZC}(\bm{Y}_g) \\
    \end{array}
\end{equation}
This design avoids conflicts between parallel branches during the early training stage,
while enabling progressive development of information alignment paths.

\begin{table*}[t]
    \centering
    \caption{Quantitative comparisons with different base models demonstrate that
        our module directly lifts different base multi-view diffusion models for 3D asset generation.
        MVD-v1.5-3D, MVD-v2.1-3D, and DV-3D denote MVDream 1.5 with \moduleName, MVDream 2.1 with \moduleName,
        and Dreamview with \moduleName, respectively.
        '*' indicates that we omit the FID for DiffSplat because
        this approach is trained on the complete G-Objaverse dataset and thus our evaluation subset is covered by their training dataset.
        $^\dagger$ indicates that we leverage TexGaussian to generate PBR materials to calculate the $\mathrm{FID}_{mat}$.}
        \label{tab:quant_t23d}
    \begin{tabular}{c|cccccccc}
        \toprule
        \toprule
         & LaRa & LGM & DiffSplat & TRELLIS$^\dagger$ & 3DTopia-XL & MVD-v1.5-3D & MVD-v2.1-3D & DV-3D \\
        \midrule
        \midrule
        FID$\downarrow$ & 38.80 & 36.00 & -* & 27.08 & 64.96 & 40.76 & 36.38 & 33.75 \\
        $\mathrm{FID}_{mat}$$\downarrow$ & - & - & - & 44.05 & 47.19 & 53.62 & 41.47 & 42.63 \\
        IS$\uparrow$ & $12.77 \pm 0.39$ & $13.53 \pm 0.52$ & $13.75 \pm 0.51$ & $13.40 \pm 0.28$ & $9.42 \pm 0.12$ & $12.88 \pm 0.46$ & $14.63 \pm 0.51$ & $13.91 \pm 0.38$ \\
        CLIP score$\uparrow$ & 31.41 & 32.17 & 31.62 & 31.25 & 29.26 & 32.84 & 33.70 & 34.36 \\
        \bottomrule
        \bottomrule
    \end{tabular}
\end{table*}

\begin{figure*}[t]
  \begin{overpic}[width=0.99\linewidth]{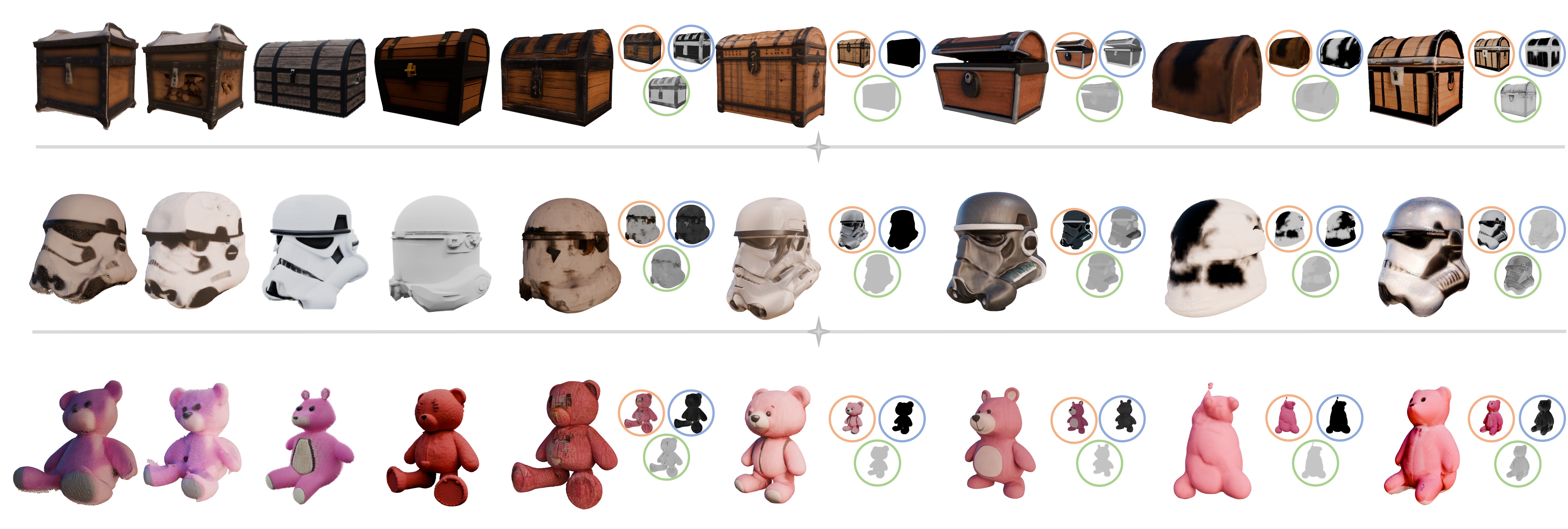}
  \put(32, 33){``\textit{A wooden chest with a lock and black trim, 3d asset.}''}
  \put(35, 22){``\textit{Star Wars Stormtrooper helmet, 3d asset.}''}
  \put(31, 10.3){``\textit{A pink teddy bear with a zipper on its back, 3d asset.}''}
  \put(3.5, -1){LaRa}
  \put(10, -1){LGM}
  \put(16, -1){DiffSplat}
  \put(24, -1){TRELLIS}
  \put(35.5, -1){T + TexG}
  \put(47, -1){Meshy-4*}
  \put(60.5, -1){Luma Genie*}
  \put(75, -1){3DTopia-XL}
  \put(90.5, -1){Ours}
  
  \put(89, 32.7){Albedo}
  \put(96, 32.7){Metallic}
  \put(91.5, 24.8){Roughness}
  \end{overpic}
  \vspace{2pt}
  \caption{
  Visual comparisons of text-conditioned 3D asset generation methods.
  For LGM and LaRa, we use MVDream 2.1 to generate four input views.
  '*' refers to the non-publicly available commercial software. 
  'T + TexG' refers to the two stage pipeline where geometry is generated via TRELLIS
  and PBR materials are produced by TexGaussian.
  }
  \label{fig:t23d}
\end{figure*}

\myPara{\moduleName~Decoder}
We design the \moduleName~based on latent diffusion models,
which produce latent feature maps at a lower resolution, resulting in a limited number of Gaussian primitives 
insufficient for representing fine-grained structures.
To address this limitation, we utilize the pre-trained decoder of the variational autoencoder (VAE)
to construct our \moduleName~Decoder, as depicted in \Figref{fig:pipe}. 
By decoding to a higher spatial resolution, our approach enables the generation 
of a greater number of Gaussian primitives, thus capturing more detailed geometric and appearance information.
In practice, we freeze most layers of the pre-trained decoder to stabilize the training,
unlock the input layers, and adjust the output layers to produce Gaussian splat attributes.
The Gaussian head decodes pixel-aligned Gaussian splat images,
where each pixel is a Gaussian primitive with 3D position $\bm{\mu} \in \mathbb{R}^3$, rotation quaternion $\bm{q} \in \mathbb{R}^4$,
scaling vector $\bm{s} \in \mathbb{R}^2$, opacity $o \in \mathbb{R}$, color $\bm{c} \in \mathbb{R}^3$, albedo $\bm{a} \in \mathbb{R}^3$,
metallic $m \in \mathbb{R}$, and roughness $r \in \mathbb{R}$.

\myPara{Integration of \moduleName~with multiple MV diffusion models}
As illustrated in \Figref{fig:pipe}, our framework is capable of combining multiple 
diffusion priors within a unified model, thus effectively harnessing the complementary strengths of different base models.
To validate the efficacy of this design, we perform extensive experiments 
on the challenging task of end-to-end 3D asset generation with PBR materials. 
We integrate robust multi-view RGB priors such as 
MVDream~\cite{shi2023mvdream}, ImageDream~\cite{wang2023imagedream}, and DreamView~\cite{yan2024dreamview}, 
alongside the multi-view PBR prior IDArb~\cite{li2024idarb}, showcasing the versatility and superior performance of our integrated framework.
For text-conditioned generation models, we construct the \moduleName~Wrapper layers with the middle and upsampling blocks 
from both text-to-MV diffusion models and the IDArb model to form a parallel branch structure, then we leverage the \moduleName~Switcher
to enable information alignment between different branches, ultimately predicting the Gaussian splat assets with the \moduleName~Decoder.
For image-conditioned generation models, layers from text-conditioned diffusion models encapsulated in the \moduleName~Wrapper 
can be straightforwardly replaced by those from image-conditioned diffusion models.

\begin{table*}[t]
    \centering
    \caption{Quantitative metrics for image-conditioned generation results. $^\dagger$ indicates that 
    we leverage TexGaussian to generate PBR materials to calculate the $\mathrm{PSNR}_{albedo}$, $\mathrm{MSE}_{metallic}$ and $\mathrm{MSE}_{roughness}$.}
    \label{tab:quant_i23d}
    \begin{tabular}{c|cccccc|cc}
        \toprule
        \toprule
         & PSNR$\uparrow$ & SSIM$\uparrow$ & LPIPS$\downarrow$ & $\mathrm{PSNR}_{albedo}\uparrow$ & $\mathrm{MSE}_{metallic}\downarrow$ & $\mathrm{MSE}_{roughness}\downarrow$ & CD$\downarrow$ & F-score@0.1$\uparrow$ \\
        \midrule
        \midrule
        LaRa & 13.32 & 0.750 & 0.344 & - & - & - & 0.075 & 72.00 \\
        LGM & 15.98 & 0.781 & 0.249 & - & - & - & 0.061 & 79.05 \\
        TRELLIS$^\dagger$ & 14.64 & 0.761 & 0.270 & 11.74 & 0.118 & 0.061 & 0.039 & 86.78 \\
        3DTopia-XL & 13.36 & 0.756 & 0.334 & 13.42 & 0.122 & 0.037 & 0.093 & 59.20 \\
        ImageDream $w/$ \moduleName & 17.04 & 0.788 & 0.227 & 17.62 & 0.049 & 0.015 & 0.057 & 80.56 \\
        \bottomrule
        \bottomrule
    \end{tabular}
\end{table*}

\begin{figure*}[t]
  \begin{overpic}[width=0.99\linewidth]{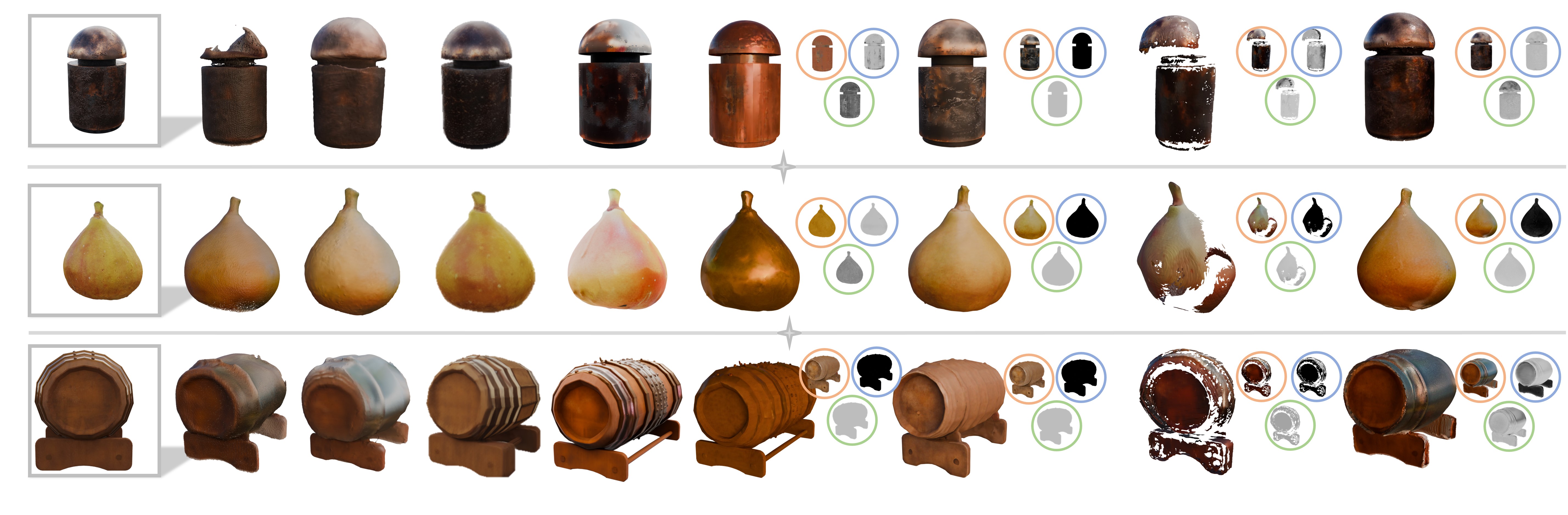}
  \put(3.5, -1){Input}
  \put(12.7, -1){LaRa}
  \put(20, -1){LGM}
  \put(26.5, -1){DiffSplat}
  \put(36, -1){TRELLIS}
  \put(45, -1){T + TexG}
  \put(58, -1){Meshy-4*}
  \put(74, -1){3DTopia-XL}
  \put(91.5, -1){Ours}
  
  \put(89, 32.2){Albedo}
  \put(96, 32.2){Metallic}
  \put(91.5, 23.7){Roughness}
  \end{overpic}
  \vspace{2pt}
  \caption{
  Visual comparisons of image-conditioned 3D asset generation methods.
  For LGM and LaRa, we use ImageDream to generate four input views.
  '*' refers to the non-publicly available commercial software.
  'T + TexG' refers to the two stage pipeline where geometry is generated via TRELLIS
  and PBR materials are produced by TexGaussian.}
  \label{fig:i23d}
\end{figure*}

\subsection{Deferred shading loss}
\label{subsec:pbr_rend}

Although we have direct access to the G-buffer information for supervision, the simultaneous generation 
of geometry and appearance for 3D assets remains a highly ill-posed problem. 
Inspired by \cite{munkberg2022extracting,liang2024gs},
we leverage an image-based differentiable rendering approach to link the rendered G-buffer information with the final RGB appearance
to reduce the ambiguity.
Specifically, we separate the rendering equation~\cite{kajiya1986rendering} as the diffuse ($\bm{L}_d$) and specular ($\bm{L}_s$) components:
\begin{equation}
    \bm{L}_o(\bm{x}, \omega_o) = \bm{L}_d + \bm{L}_s
\end{equation}
where $\bm{L}_o$ models the outgoing radiance of the surface point $\bm{x}$ at direction $\omega_o$.

For the diffuse $\bm{L}_d$ component, we employ an image-based lighting model and the split-sum approximation 
for direct illumination, while occlusion and indirect illumination are precomputed and stored into view-dependent images $\bm{I}_o$ and $\bm{I}_{irr}$:
\begin{equation}
    \bm{L}_d \approx (1 - \bm{I}_o)\int_{\Omega}\bm{L}_i(\bm{x}, \omega_i)(\omega_i, \bm{n})d\omega_i + 
    \bm{I}_o\bm{I}_{irr}
\end{equation}
where $\Omega$ denotes the upper hemisphere.
The first term models the direct illumination 
depending on the parameters $cos\theta=\omega_i\cdot\mathbf{n}$ and the roughness $r$,
which can be precomputed into a lookup map.
The second term encapsulates indirect illumination, recovered for each training view through efficient preprocessing.

The specular term $\bm{L}_s$ follows the split-sum approximation from~\cite{munkberg2022extracting}:
\begin{equation}
    \bm{L}_s \approx \int_{\Omega}\frac{DFG}{4(\bm{n}, \omega_i)(\bm{n}, \omega_o)}d\omega_i\int_{\Omega}D\bm{L}_i(\omega_i)(\bm{n}, \omega_i)d\omega_i
\end{equation}
where the first integral term models the specular BSDF under white light, 
and the second term integrates the incoming radiance using a pre-integrated cubemap.

\subsection{Geometry and Texture Refinement}
\label{subsec:post}

To improve the usability of the generated Gaussian assets, 
we introduce a dedicated post-processing procedure to convert the Gaussian assets into PBR-ready mesh assets.
Specifically,
we first extract meshes from 2DGS via TSDF fusion
and refine them through continuous remeshing~\cite{palfinger2022continuous} to acquire watertight meshes.
Next, we initialize texture maps for the 3D assets using the Gaussian assets and 
leverage a differentiable renderer to further refine the PBR materials.

\myPara{Geometry extraction and refinement}
To extract meshes from 2DGS, we render albedo and depth images along circular camera paths 
at elevations of $[10\degree, 15\degree, 20\degree]$ 
around each instance, along with additional top and bottom views,
and we utilize the ScalableTSDFVolume from Open3D~\cite{Zhou2018} with a voxel size of 0.008 and 
a truncation threshold of 0.02 to perform TSDF Fusion to extract the initial mesh.
Then, we compute the convex hull of the initial mesh to fill all the holes in the original mesh.
Finally, we render normal maps and alpha maps from the 2DGS around the instances as the target views
and perform 100 iterations of continuous remeshing~\cite{palfinger2022continuous} 
to transform the convex hull into smooth, high-quality watertight meshes.
This process improves the robustness and watertightness of mesh extraction, but 
may oversmooth structures with thin or high-genus topology. We provide detailed visualizations and analysis 
in the supplementary materials.

\myPara{Texture initialization and refinement}
After mesh extraction, we generate UV maps using Blender's Smart UV Project~\cite{blender}.
The rendered albedo, metallic, and roughness maps from the 2DGS are then unprojected onto the mesh to initialize the PBR materials.
Then, following  \cite{Ummenhofer2024OWL}, we use the differentiable 
renderer~\cite{jakob2022mitsuba3,jakob2022dr}
to align four orthogonal views with the images generated by multi-view diffusion models, 
thereby enhancing the visual quality of the final appearance. 
The geometry refinement process may simplify the topology;
for example, some hollow regions can be sealed into closed surfaces, 
which may introduce unreliable areas for texture refinement.
To reduce its effect, we adopt two robustness rules.
First, for newly created surfaces (e.g., due to hole filling) and unseen regions,
we initialize the canonical PBR parameters to a conservative prior: 
albedo $[0.0, 0.0, 0.0]$, metallic $0.0$, and roughness $1.0$, so that these regions do not spuriously attract specular highlights or hallucinated textures.
Second, we render alpha masks from the predicted 2DGS and restrict texture refinement to the masked valid regions,
since the 2DGS provides more reliable visibility cues for thin or open structures.
The entire procedure takes less than 30 seconds on an NVIDIA GeForce RTX 4090 GPU.

\subsection{Training and Inference}
\label{subsec:train_infer}

\begin{figure*}
  \begin{overpic}[width=0.99\linewidth]{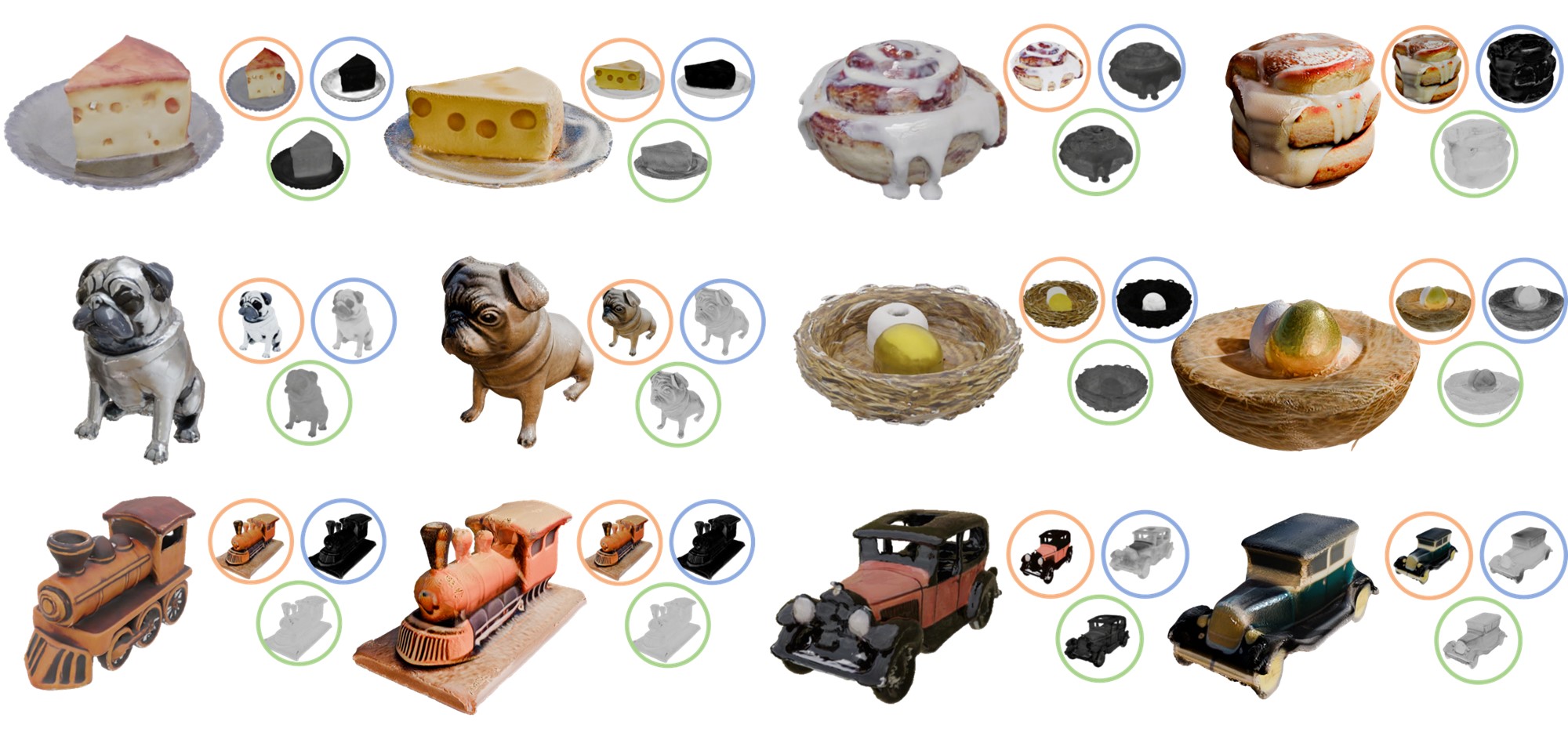}
  \put(11, 46){‘’\textit{A wedge of cheese on a silver platter.}‘’}
  \put(59, 46){‘’\textit{A fresh cinnamon roll covered in glaze.}‘’}
  \put(15, 31.5){‘’\textit{A pug made out of metal.}‘’}
  \put(57, 31.5){‘’\textit{A nest with white eggs and one golden egg.}‘’}
  \put(12.7, 16){‘’\textit{A train engine made out of clay.}‘’}
  \put(67.5, 16){‘’\textit{An old vintage car.}‘’}
  \put(87.5, 15.3){Albedo}
  \put(94, 15.3){Metallic}
  \put(90, 2.2){Roughness}

  \put(5, 0){Meta 3D AssetGen}
  \put(35, 0){Ours}
  \put(56, 0){Meta 3D AssetGen}
  \put(84, 0){Ours}
  \end{overpic}
  \caption{
  We provide detailed visual comparisons with Meta 3D AssetGen using the same prompts in \cite{siddiqui2024meta}.
  Our method is capable of producing PBR-ready 3D assets.
    For example in the cheese case, our method clearly separates the metal platter from the cheese.}
  \label{fig:meta}
\end{figure*}

\myPara{Training}
During training, we sample batches consisting of four orthogonal views and four random views, 
each comprising RGB, alpha mask, normal, depth, albedo, metallic, and roughness maps, all scaled at a resolution of 256.
We train our model using bfloat16 precision and a gradient accumulation step of 8, 
with each GPU processing two batches, resulting in a total batch size of 128.
We add random grid distortion~\cite{tang2024lgm} to the four orthogonal views,
assign random background color to the RGB images,
and then process the views following MVDream, ImageDream, and Dreamview
with random noise level $t\in[0, 1000)$ to get the input for the 2D diffusion models.
Then, we render RGB, albedo, alpha mask, metallic, roughness, depth, and normal maps for all eight views at the same resolution.
For RGB and albedo supervision, 
we apply MSE loss, SSIM loss~\cite{kerbl3Dgaussians}, and LPIPS loss~\cite{zhang2018unreasonable}:
\begin{equation}
    \begin{array}{ccc}
    \mathcal{L}_{color} & = & \lambda_1 \mathcal{L}_{MSE}(\bm{I}_{color}, \bm{I}_{color}^{GT}) + \\
     & & \lambda_2 \mathcal{L}_{SSIM}(\bm{I}_{color}, \bm{I}_{color}^{GT}) + \\
     & & \lambda_3 \mathcal{L}_{LPIPS}(\bm{I}_{color}, \bm{I}_{color}^{GT})
    \end{array}
    \label{equ:rgb}
\end{equation}
where $\lambda_1=1, \lambda_2=2, \lambda_3=5$. For alpha map, we use binary cross-entropy loss:
\begin{equation}
    \mathcal{L}_{alpha} = \mathcal{L}_{BCE}(\bm{I}_{alpha}, \bm{I}_{alpha}^{GT})
\end{equation}
For metallic and roughness, we apply MSE loss:
\begin{equation}
    \mathcal{L}_{material} = \mathcal{L}_{MSE}(\bm{I}_{material}, \bm{I}_{material}^{GT})
\end{equation}
And we also apply the depth distortion loss and normal consistency loss~\cite{huang20242d}.
The depth distortion loss is applied with a weight of $2 \times 10^4$ during the initial three epochs 
to facilitate convergence, subsequently decaying to $1 \times 10^2$.
\myPara{Fine-tuning with deferred shading loss}
After 20 epochs of training with direct supervision using G-buffer information, 
we conduct further fine-tuning steps on a dataset subset using the deferred shading loss described in \Secref{subsec:pbr_rend}.
During this phase, all original training losses are retained, and image-based rendering is employed to align the albedo, 
metallic, roughness, and normal maps with $\mathbf{I}_{shade}$.
The shaded images are then supervised using the loss defined in~\eqref{equ:rgb}.
This additional fine-tuning stage encompasses another 20 epochs.
The training and fine-tuning span 3 days on 1 node of 8 NVIDIA H20 GPUs.

\myPara{Inference}
We infer our model via DDIM sampling with 50 steps and a guidance scale of 7.5 
for the MVDream-based and DreamView-based version.
For the ImageDream-based model, we use a guidance scale of 5.0 as in \cite{wang2023imagedream}.
Since our \moduleName~receives feature maps from diffusion models and generates Gaussian splats in a feed-forward manner, 
only a single inference pass with our model is required during the sampling process of the multi-view diffusion models.
Empirically, we find that utilizing feature maps from diffusion models at a noise level of $t=150$ yields optimal results.



\begin{figure*}[t]
  \centering
  \begin{overpic}[width=0.99\linewidth]{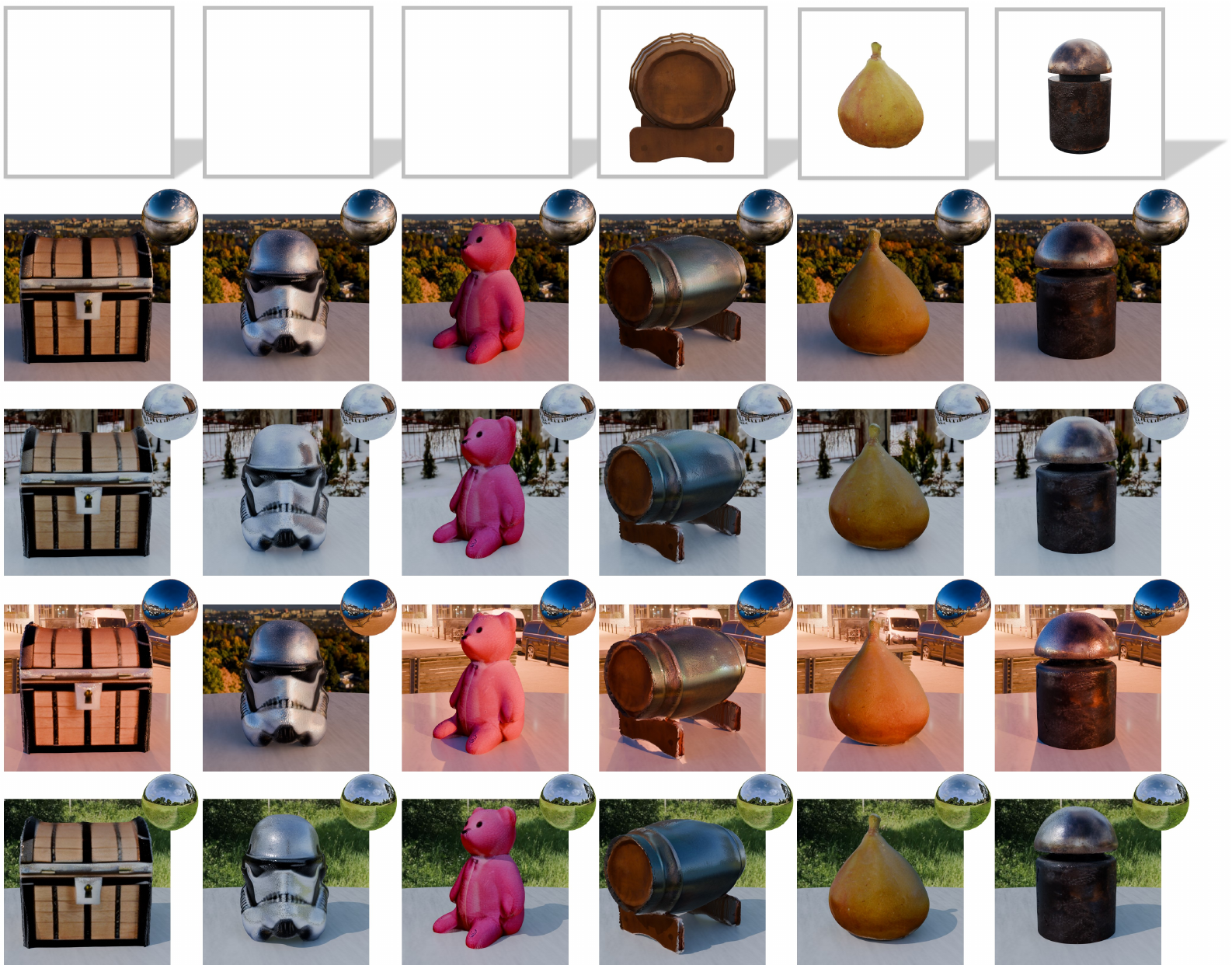}
  \put(1.4, 75.5){\textbf{Input prompt}}
  \put(1.0, 73){''\textit{A wooden chest}}
  \put(1.4, 70.5){\textit{with a lock and}}
  \put(1.4, 68){\textit{black trim, 3d}}
  \put(1.4, 65.5){\textit{asset.}''}

  \put(17.5, 75.5){\textbf{Input prompt}}
  \put(17.5, 73){\textit{''Star Wars}}
  \put(17.5, 70.5){\textit{Stormtrooper}}
  \put(17.5, 68){\textit{helmet, 3d}}
  \put(17.5, 65.5){\textit{asset.''}}

  \put(33.7, 75.5){\textbf{Input prompt}}
  \put(33.7, 73){\textit{''A pink teddy}}
  \put(33.7, 70.5){\textit{bear with a}}
  \put(33.7, 68){\textit{zipper on its}}
  \put(33.7, 65.5){\textit{back, 3d asset.''}}

  \put(50.2, 75.7){\textbf{Input image}}
  \put(66.3, 75.7){\textbf{Input image}}
  \put(82.5, 75.7){\textbf{Input image}}
  \end{overpic}
  \caption{
  Relighting results under different HDRI maps.
  The synthesized diffuse materials exhibit base color changes under different environment maps, as shown in the fruit and teddy examples,
  while specular materials, as the barrel and helmet instances, correctly reflect the surroundings.
  }
  \label{fig:relit}
\end{figure*}

\section{Experiments}
\label{sec:exp}

\subsection{Experimental settings}
\label{subsec:exp_sets}

\myPara{Dataset}
We conduct all experiments using the open-source G-Objaverse dataset~\cite{qiu2024richdreamer},
a high-quality multi-view dataset that includes G-buffer information.
This dataset comprises 265K rendered 3D instances from Objaverse~\cite{deitke2023objaverse} and 779K 3D instances 
from the Objaverse-XL Alignment~\cite{deitke2024objaverse}, each instance featuring 38 views around the object.
We filter this dataset based on aesthetic scores, material quality, and valid pixel coverage, resulting in a high-quality subset of 69k instances for training.
Within this subset, we apply the deferred shading function (described in \Secref{subsec:pbr_rend}) 
to ground truth (GT) G-buffer maps to recover environment maps, as well as view-dependent occlusion and indirect illumination maps, for fine-tuning purposes.
We compute the PSNR between the shaded images, shaded with GT G-buffer maps and recovered lighting information, and the GT images, 
selecting instances with a PSNR greater than 35 dB for fine-tuning.
This results in a subset of 10K high-quality instances with recovered environmental and illumination maps.

\begin{figure*}[t]
  \subfloat[Detailed visualizations of the more generated 3D assets with text conditions.]{
      \begin{overpic}[width=0.99\linewidth]{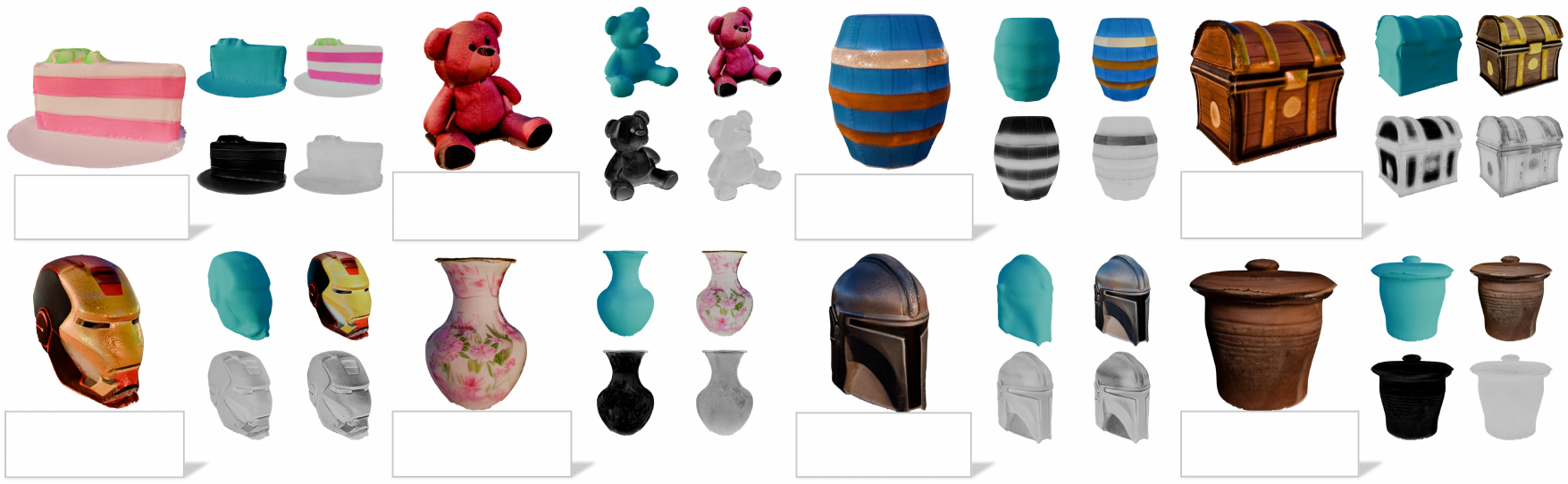}
      \scriptsize
        \put(13, 23){geometry}
        \put(19.7, 23){albedo}
        \put(13, 17){metallic}
        \put(18.6, 17){roughness}

        \put(1.3, 18){\textit{A slice of cake on}}
        \put(1.3, 16.5){\textit{a plate.}}
        \put(25.5, 18.7){\textit{A pink teddy bear}}
        \put(25.5, 17.2){\textit{with zipper on its}}
        \put(25.5, 15.7){\textit{back.}}
        \put(51.5, 18){\textit{A blue and golden}}
        \put(51.5, 16.5){\textit{wooden barrel.}}
        \put(75.8, 18){\textit{A wooden chest }}
        \put(75.8, 16.5){\textit{with gold coins.}}

        \put(0.7, 3){\textit{A red and yellow }}
        \put(0.7, 1){\textit{Iron Man helmet.}}
        \put(25.5, 3){\textit{A large pink and}}
        \put(25.5, 1){\textit{white vase.}}
        \put(51.3, 3){\textit{A silver Manda-}}
        \put(51.3, 1){\textit{lorian helmet.}}
        \put(75.8, 3){\textit{A large clay jar }}
        \put(75.8, 1){\textit{with a lid.}}
      \end{overpic}
  }
  \vfill
  \subfloat[Detailed visualizations of the more generated 3D assets with image conditions.]{
      \begin{overpic}[width=0.99\linewidth]{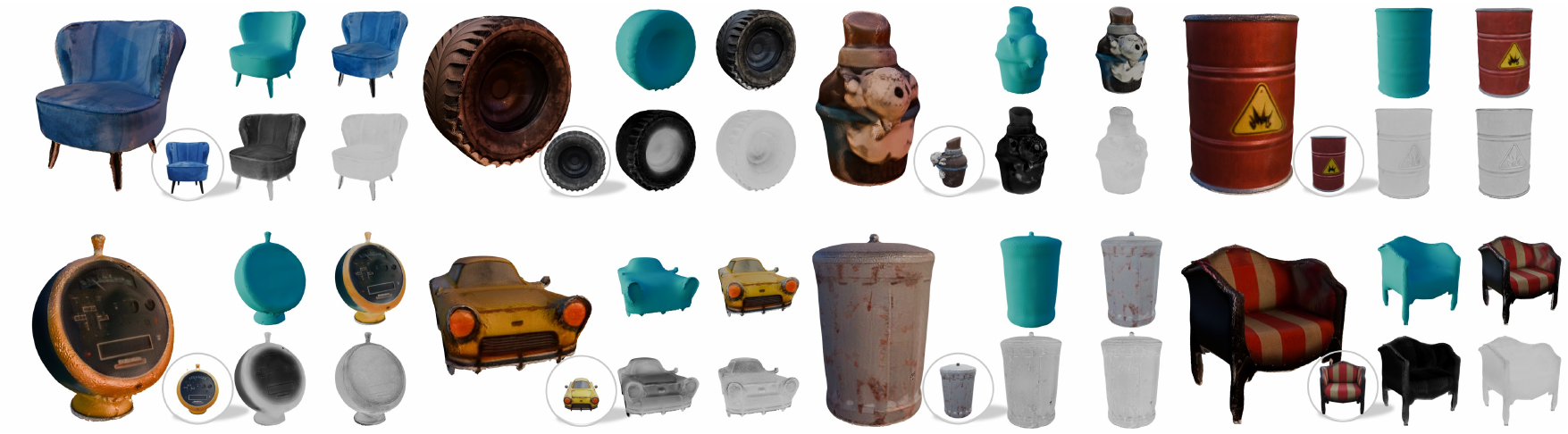}
      \scriptsize
        \put(13.8, 20.5){geometry}
        \put(21.5, 20.5){albedo}
        \put(14.2, 14.5){metallic}
        \put(20.2, 14.5){roughness}

        \put(10.5, 14){\textit{input}}
        \put(10.5, 12.5){\textit{image}}
      \end{overpic}
  }
  \caption{We provide detailed visualizations of geometry and PBR materials from the generated 3D assets, along with
  the input conditions. The results indicate that the proposed \moduleName~is capable of generating a diversity of high-quality PBR-ready 3D assets with either text conditions
  or image conditions.}
  \label{fig:batch_res}
\end{figure*}

\myPara{Baselines}
We select MVDream 1.5, MVDream 2.1~\cite{shi2023mvdream}, and DreamView~\cite{yan2024dreamview} 
as text-to-3D baselines, and ImageDream~\cite{wang2023imagedream} as the image-to-3D baseline for our experiments.
Additionally, we incorporate knowledge from IDArb~\cite{li2024idarb} into our framework to enhance PBR material generation.
For evaluating 3D asset generation, we compare against 3DTopia-XL~\cite{chen20243dtopia}, 
a baseline supporting text- and image-conditioned 3D asset generation with PBR materials.
We also compare our results with DiffSplat~\cite{lin2025diffsplat}, LGM~\cite{tang2024lgm}, and LaRa~\cite{chen2024lara}. 
DiffSplat~\cite{lin2025diffsplat} supports native 3D Gaussian splat generation, whereas LGM and LaRa are reconstruction-based Gaussian generation methods.
Furthermore, we report evaluation metrics against TRELLIS~\cite{xiang2025structured}, 
a leading native 3D generation method supporting text- and image-conditioned Gaussian splat generation. 
We also introduce a two-stage pipeline for comparison,
where we apply TexGaussian~\cite{xiong2025texgaussian}
to the generated meshes from TRELLIS to predict 
PBR materials,
which forms a representative baseline that 
decouples the task into geometry generation 
and PBR mateiral generation.
Besides, we also compare our results against Meshy-v4~\cite{meshy} and LumaAI-Genie~\cite{genie},
proprietary software solutions for 3D mesh generation with PBR materials.

\myPara{Text-conditioned evaluation metrics}
Following the evaluation protocol outlined by MVDream~\cite{shi2023mvdream}, 
we randomly select 1,000 unseen prompts from the G-Objaverse dataset, with each instance 
represented by 12 evenly distributed views that serve as ground truth images.
We evaluate the alignment between rendered images and text prompts using the CLIP similarity score~\cite{radford2021learning}.
Furthermore, we employ the Fréchet Inception Distance (FID)~\cite{heusel2017gans} and Inception Score (IS)~\cite{salimans2016improved} 
to assess the quality of rendered images from the generated 3D assets.
To evaluate the quality of PBR materials, we render 12 views of albedo, metallic, and roughness maps. 
The metallic ($m$) and roughness ($r$) maps are stored in the format $(o, r, m)$, with the $o$ channel unused.
Finally, we report the $\mathrm{FID}_{mat}$ metric, which compares the generated material maps against ground truth material maps.

\myPara{Image-conditioned evaluation metrics}
We also perform inference on the same 1,000 unseen instances from the text-conditioned evaluation,
selecting 12 views per instance as ground truth images, with the front view serving as the input view.
We report the PSNR, SSIM~\cite{wang2004image}, and LPIPS~\cite{zhang2018unreasonable} metrics 
to evaluate the rendered images.
For the evaluation of PBR materials, we report PSNR for albedo maps
and report MSE for the metallic and roughness images.
To evaluate the geometry quality, we align the generated meshes 
with ground truth meshes via the ICP algorithm implemented in Open3D~\cite{Zhou2018},
re-scale all meshes into a cube of size $[-0.5, 0.5]^3$ as \cite{qiu2024richdreamer,xiang2025structured},
and report the Chamfer Distance (CD) and F-Score with a threshold of 0.1.

\subsection{Main results}
\label{subsec:gen}

\myPara{Text-conditioned 3D asset generation}
As shown in \tabref{tab:quant_t23d}, 
the proposed \moduleName~demonstrates compatibility with different base models.
Most importantly, the generation quality improves as the capacity of the base models increases, as evidenced by the FID and CLIP scores
of \moduleName~with different base models.
This indicates that our scheme can effectively exploit priors from base models for 3D generation, demonstrating its potential.
Furthermore, our models outperform the state-of-the-art method 3DTopia-XL~\cite{chen20243dtopia} on the challenging task of end-to-end PBR-ready 3D asset generation.
Notably, 3DTopia-XL is trained on 16 nodes with 8 NVIDIA A100 GPUs each and 256k 3D instances, 
whereas our \moduleName~is trained on a single node with 8 NVIDIA H20 GPUs and only 69k multi-view images
to lift pre-trained MV diffuison models for 3D generation, 
which demonstrates the efficiency and effectiveness of our scheme.
Although our method achieves a higher FID compared to the native 3D generation approach TRELLIS~\cite{xiang2025structured}—primarily 
because our task requires the simultaneous synthesis of geometry and PBR materials, while TRELLIS focuses only on geometry and simple vertex colors—
our approach achieves a better CLIP score than TRELLIS, indicating better alignment between the generation results and the prompts.
Lastly, our method outperforms the two-stage pipeline on PBR material generation,
which confirms the strength our scheme.

\begin{table*}[t]
    \centering
    \caption{Quantitative evaluation of the ablation study experiments.}
    \label{tab:quant_abla}
    \subfloat[Ablation study on the text-conditioned model.]{
    \centering
    \begin{tabular}{c|cccccc}
        \toprule
        \toprule
         & $w/o$ \moduleName-W & $w/o$ \moduleName-S & $w/o$ \moduleName-D & $w/o$ IDArb & $w/o$ deferred shading & Full model \\
        \midrule
        \midrule
        FID$\downarrow$ & 54.52 & 38.33 & 37.39 & 40.89 & 36.73 & 36.38 \\
        $\mathrm{FID}_{mat}$$\downarrow$ & 74.31 & 47.48 & 42.37 & 51.72 & 43.52 & 41.47 \\
        IS$\uparrow$ & $13.10 \pm 0.54$ & $14.46 \pm 0.63$ & $14.96 \pm 0.47$ & $14.52 \pm 0.56$ & $14.84 \pm 0.57$ & $14.63 \pm 0.51$ \\
        CLIP score$\uparrow$ & 32.76 & 33.60 & 33.73 & 33.40 & 33.73 & 33.70 \\
        \bottomrule
        \bottomrule
    \end{tabular}
    \label{tab:abla_naive_txt}
    }
    \vfill
    \subfloat[Ablation study on the image-conditioned model.]{
    \centering
    \begin{tabular}{c|cccccc}
        \toprule
        \toprule
         & $w/o$ \moduleName-W & $w/o$ \moduleName-S & $w/o$ \moduleName-D & $w/o$ IDArb & $w/o$ deferred shading & Full model \\
        \midrule
        \midrule
        PSNR$\uparrow$ & 14.52 & 16.47 & 16.86 & 16.43 & 17.03 & 17.04 \\
        SSIM$\uparrow$ & 0.698 & 0.774 & 0.786 & 0.768 & 0.788 & 0.788 \\
        LPIPS$\downarrow$ & 0.335 & 0.247 & 0.231 & 0.252 & 0.227 & 0.227 \\
        $\mathrm{PSNR}_{albedo}\uparrow$ & 14.66 & 16.90 & 17.42 & 16.79 & 17.60 & 17.62 \\
        $\mathrm{MSE}_{metallic}\downarrow$ & 0.079 & 0.072 & 0.056 & 0.077 & 0.053 & 0.049 \\
        $\mathrm{MSE}_{roughness}\downarrow$ & 0.022 & 0.020 & 0.016 & 0.021 & 0.015 & 0.015 \\
        \bottomrule
        \bottomrule
    \end{tabular}
    \label{tab:abla_naive_img}
    }
\end{table*}

\begin{figure*}[t]
  \centering
  \begin{overpic}[width=0.9\linewidth]{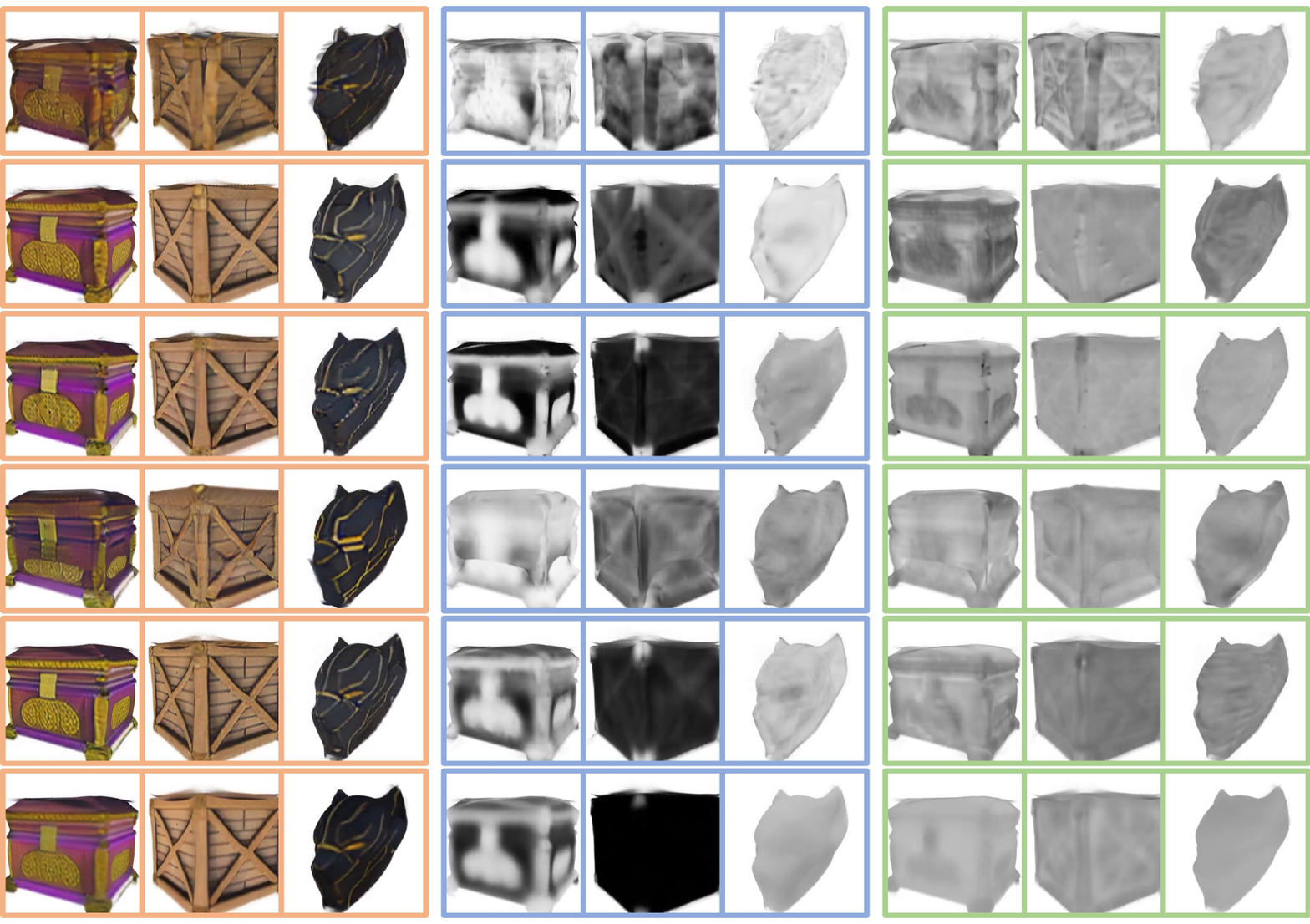}
  \put(12, -1.7){Albedo}
  \put(46, -1.7){Metallic}
  \put(78, -1.7){Roughness}
  {\footnotesize
  \put(-2, 59){\rotatebox{90}{$w/o$ \moduleName-W}}
  \put(-2, 48){\rotatebox{90}{$w/o$ \moduleName-S}}
  \put(-2, 36){\rotatebox{90}{$w/o$ \moduleName-D}}
  \put(-2, 25){\rotatebox{90}{$w/o$ IDArb}}
  \put(-2, 15){\rotatebox{90}{$w/o$ DS}}
  \put(-2, 2){\rotatebox{90}{Full model}}
  }
  \end{overpic}
  \vspace{4pt}
  \caption{
  Visual comparisons from the ablation studies. As our LGAA directly produces 2DGS with PBR channels,
  we visualize the renderings from the produced 2DGS to exhibit the impact of different modules.
  'DS' denotes deferred shading.
  The three instances, from left to right, correspond to the descriptions: ''a black panther mask in the shape of a cat head'', ''an ornate wooden box with a lock'', 
  and ''a wooden crate''. The results highlight the contributions of each component to our framework.
  }
  \label{fig:abla}
\end{figure*}

Qualitative comparisons are illustrated in \Figref{fig:t23d}.
Our approach generates accurate geometry and fine-grained PBR maps across different materials.
In contrast, 3DTopia-XL produces inaccurate geometry or blurry appearances, 
highlighting the robustness and superiority of our approach.
Furthermore, TRELLIS does not support the decoupling of PBR materials,
and it suffers from out-of-domain issues, as shown in the second row of \Figref{fig:t23d}.
In addition, other GS-based generation or reconstruction methods fail to produce relightable 3D assets.
For example, LGM and LaRa reconstruct coarse geometry with appearances entirely baked into vertex colors, 
while DiffSplat produces Gaussian splats that only support novel view synthesis.

\myPara{Image-conditioned 3D asset generation}
As shown in \tabref{tab:quant_i23d} and \Figref{fig:i23d}, our module achieves superior results both quantitatively and qualitatively,
demonstrating the compatibility of our scheme with image-conditioned multi-view diffusion models.
Our method not only faithfully generates complete geometry but also synthesizes accurate PBR materials for various objects, 
ensuring photorealistic rendering and relighting.
In contrast, the state-of-the-art 3DTopia-XL method even struggles to generate accurate geometry, 
whereas our approach successfully separates different materials with correct PBR attributes and recovers accurate geometry for specular instances.
Additionally, TRELLIS bakes PBR materials into simple vertex colors, which tend to be oversaturated, leading to inferior quantitative metrics.
The two-stage pipeline that combines TRELLIS with TexGaussian method fails to generate accurate PBR materials,
as shown in \Figref{fig:i23d}.
For the reconstruction method LaRa, specular regions significantly degrade geometry quality,
while LGM tends to reconstruct blurry vertex colors.
The Chamfer Distance and F-Score confirms that our method truely lifts 2D diffusion priors for 3D generation.

As shown in \figref{fig:meta}, we additionally provide qualitative comparisons with a closely related method, 
Meta 3D AssetGen~\cite{siddiqui2024meta}. Since its official implementation is not publicly available, 
we restrict the comparison to qualitative results based on the figures and prompts reported in their paper.
It is worth noting that our framework efficiently lifts pretrained 2D diffusion priors to 3D 
generation in a novel scheme, which leads to efficient training:
our model can be trained on a single 8$\times$NVIDIA H20 machine, whereas Meta 3D AssetGen reports training on 64$\times$NVIDIA A100 GPUs.
This highlights the training efficiency of our approach.

\myPara{Relighting results}
We place the generated 3D assets under different HDRI maps to showcase the relighting results in \Figref{fig:relit}.
Specular materials accurately reflect the surrounding environment, as demonstrated by the helmet and barrel instances, 
while diffuse materials exhibit only changes in base color, as seen in the fruit and teddy bear example. 
These results demonstrate the accuracy of the synthesized PBR materials.
Please refer to our supplementary video for relighting results under rotating HDRI maps.

\myPara{More results}
We present detailed visualizations of the geometry and PBR materials for additional results in \Figref{fig:batch_res}, 
along with the corresponding input prompts or images.
Our text-conditioned models align the generated results well with the input prompts,
while the image-conditioned model faithfully reconstructs the input views and infers the correct geometry of the unseen parts.
More importantly, all our models are able to synthesize accurate PBR materials for various instances,
demonstrating the robustness and usability of our approach.
Additional rendering results can be found in our supplementary video.

\subsection{Ablation Study}
\label{subsec:abla}

We conduct ablation studies using the
text-conditioned MV diffusin model MVDream 2.1 and image-conditioned model ImageDream
to investigate the design choices of the proposed module.
Furthermore, given the flexibility of our approach to integrate multiple sources of knowledge within an unified framework for 3D asset generation, 
we perform an ablation study to assess the impact of incorporating IDArb priors. 
Finally, we evaluate the impact of the deferred shading loss.

\myPara{\moduleName~module designs}
We comprehensively investigate the design choices to validate the effectiveness of the proposed modules.
\moduleName-W adaptively preserves the pre-trained priors and leverages the knowledge for 3D asset generation,
serving as a critical component.
To assess the significance of this design, we unlock and train the cloned layers inside the \moduleName-W to investigate the effectiveness of this key design,
which we refer to as '$w/o$ \moduleName-W'.
Furthermore, \moduleName-D adapts the pre-trained VAE Decoder to function as a powerful upsampler for Gaussian asset generation;
therefore, we replace the pre-trained VAE Decoder layers with three layers of PixelShuffle, denoted as '$w/o$ \moduleName-D'.
Additionally, we remove other components to assess their contributions to our framework.
As shown in \tabref{tab:quant_abla} and \Figref{fig:abla}, we draw the following conclusions: 

(1) Our \moduleName~Wrapper effectively preserves knowledge and fuses priors from pre-trained models into a unified framework for end-to-end 3D asset generation.
In contrast, training the previously frozen layers increases training costs and disrupts the preserved priors, leading to degraded results,
as indicated in the '$w/o$ \moduleName-W' column of \tabref{tab:quant_abla} and the first row of \Figref{fig:abla}.
(2) Our \moduleName~Switcher plays a crucial role in aligning different priors in the unified framework.
As demonstrated in the second column of \tabref{tab:quant_abla}, the absence of the Switcher results in significant misalignment 
between the rendered RGB appearance metric (FID / PSNR) and 
the synthesized PBR metric ($\mathrm{FID}_{mat}$ / $\mathrm{PSNR}_{albedo}$ / $\mathrm{MSE}_{metallic}$ / $\mathrm{MSE}_{roughness}$).
(3) The \moduleName~Decoder serves as a powerful upsampler in latent space, 
effectively refining latent maps to generate more Gaussian primitives with fine-grained details.
As shown in the third row of \Figref{fig:abla}, replacing \moduleName-D introduces grid artifacts in the generated albedo.

\myPara{\moduleName~scaling ability}
The proposed scheme demonstrates two types of scalability.
First, it achieves improved results by incorporating more powerful base diffusion models.
As shown in \tabref{tab:quant_t23d}, \moduleName~achieves significantly better FID when paired with DreamView compared to MVDream 1.5.
Second, our approach flexibly integrates multiple diffusion priors within a unified framework for specific tasks.
As illustrated in \tabref{tab:quant_abla} and \Figref{fig:abla}, 
when integrated with the IDArb prior, our module adaptively leverages this knowledge to enhance the quality of PBR material generation.

\myPara{Deferred shading loss}
Simultaneously generating geometry and PBR materials is a highly ill-posed problem due to inherent ambiguity. 
Therefore, we introduce image-based deferred shading to couple G-buffer information with RGB appearance. 
As shown in the fifth column of \tabref{tab:quant_abla}, the deferred shading loss quantitatively improves the quality of PBR materials.
Though the quantitative improvements are marginal, the results in \Figref{fig:abla}
evidence that the deferred shading loss improves the visual quality of the PBR materials, especially in the metallic and roughness channels.


\section{Limitations}
\label{sec:limit}

Although our scheme enables the simultaneous generation of geometry and PBR materials, it has the following limitations:
First, since our approach tames pre-trained MV diffusion models for 3D asset generation by training additional adapters
while keeping the pre-trained models frozen, the dataset used to train the adapters must adhere to the 
conventions of the dataset on which the MV diffusion models were originally trained.
Second, our method is supervised solely by pixel-level losses. As a result, the internal structures of instances lack appropriate regularization. 
We leave the exploration of combining our approach with native 3D generation schemes for improved structural modeling to future work.
Third, our pipeline incorporates TSDF Fusion and continuous remeshing to achieve robust watertight mesh extraction,
which, however, may partially erase thin structures or struggle to generate high genus geometry. 
Replacing the 2DGS representation by other neural 3D representation that supports direct mesh estraction instead of TSDF Fusion 
and introduing native 3D supervision may alleviate the problem.
Lastly, the \moduleName~Sitwcher requires the MV diffusion priors to have the similar network architecture,
including the similar network layouts and feature resolutions.
How to incorporate diffusion priors with different network architecture is underexpored.


\section{Conclusion}
\label{sec:conclu}

In this paper, we explore an alternative scheme for leveraging 2D diffusion priors 
in the challenging task of end-to-end 3D asset generation with physically based rendering (PBR) materials.
We propose a novel \moduleName~that modulates pre-trained multi-view diffusion models 
to enable direct 3D asset generation, demonstrating superior generalization capability and efficiency.
Moreover, our framework exhibits scalability and flexibility by adaptively integrating 
diverse knowledge into a unified model, with its generative capabilities further enhanced when paired with more powerful base models.
Finally, we establish a complete pipeline based on the proposed method, which efficiently 
generates high-quality 3D meshes with PBR materials using consumer-grade GPUs in under 30 seconds.
Extensive quantitative and qualitative evaluations highlight the significant potential of our proposed scheme.

\section*{Acknowledgment}
This work was supported by the National Key R\&D Program of China No. 2024YFC3015801, National Science Fund of China under Grant Nos. U24A20330, 62361166670, and 62276144.

\ifCLASSOPTIONcaptionsoff
  \newpage
\fi

\bibliographystyle{IEEEtran}
\bibliography{bare_jrnl}

\newpage

\vspace{-.4in}
\begin{IEEEbiography}[{\includegraphics[width=1in]{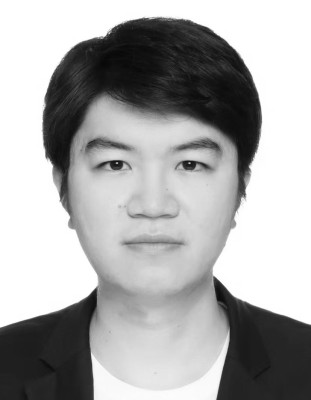}}]
{Ze-Xin Yin} received the bachelor`s degree 
from Xidian University, in 2017.
He is currently working toward the PhD degree under the supervision of 
Prof. Jin Xie 
with the College of Computer Science, Nankai University. 
His research interests mainly focus on neural radiance fields and 3D computer vision.
\end{IEEEbiography}

\vspace{-.4in}
\begin{IEEEbiography}[{\includegraphics[width=1in]{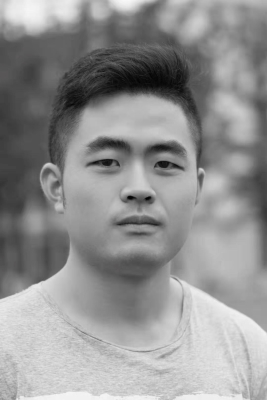}}]
{Jiaxiong Qiu} received the bachelor`s degree 
from Dalian Maritime University, in 2017, 
the master degree from the University of Electronic Science 
and Technology of China,  in 2020, 
and the PhD degree from the College of Computer Science, 
Nankai University, in 2024. 
His research interests include computer vision, computer graphics, robotics, and deep learning.
\end{IEEEbiography}

\vspace{-.4in}
\begin{IEEEbiography}[{\includegraphics[width=1in]{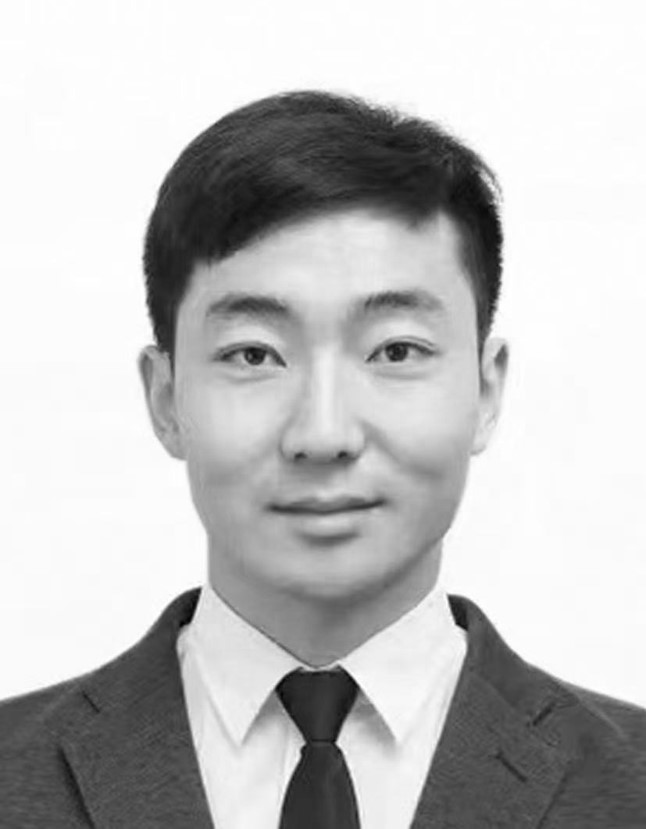}}]
{Liu Liu} is currently a 3D Vision Researcher at the Robot Lab of Horizon Robotics. 
He received his Master’s degree from Southeast University in 2019, advised by Prof. Fujun Yang. 
His research interests include 3D Vision and Embodied AI.
\end{IEEEbiography}

\vspace{-.4in}
\begin{IEEEbiography}[{\includegraphics[width=1in]{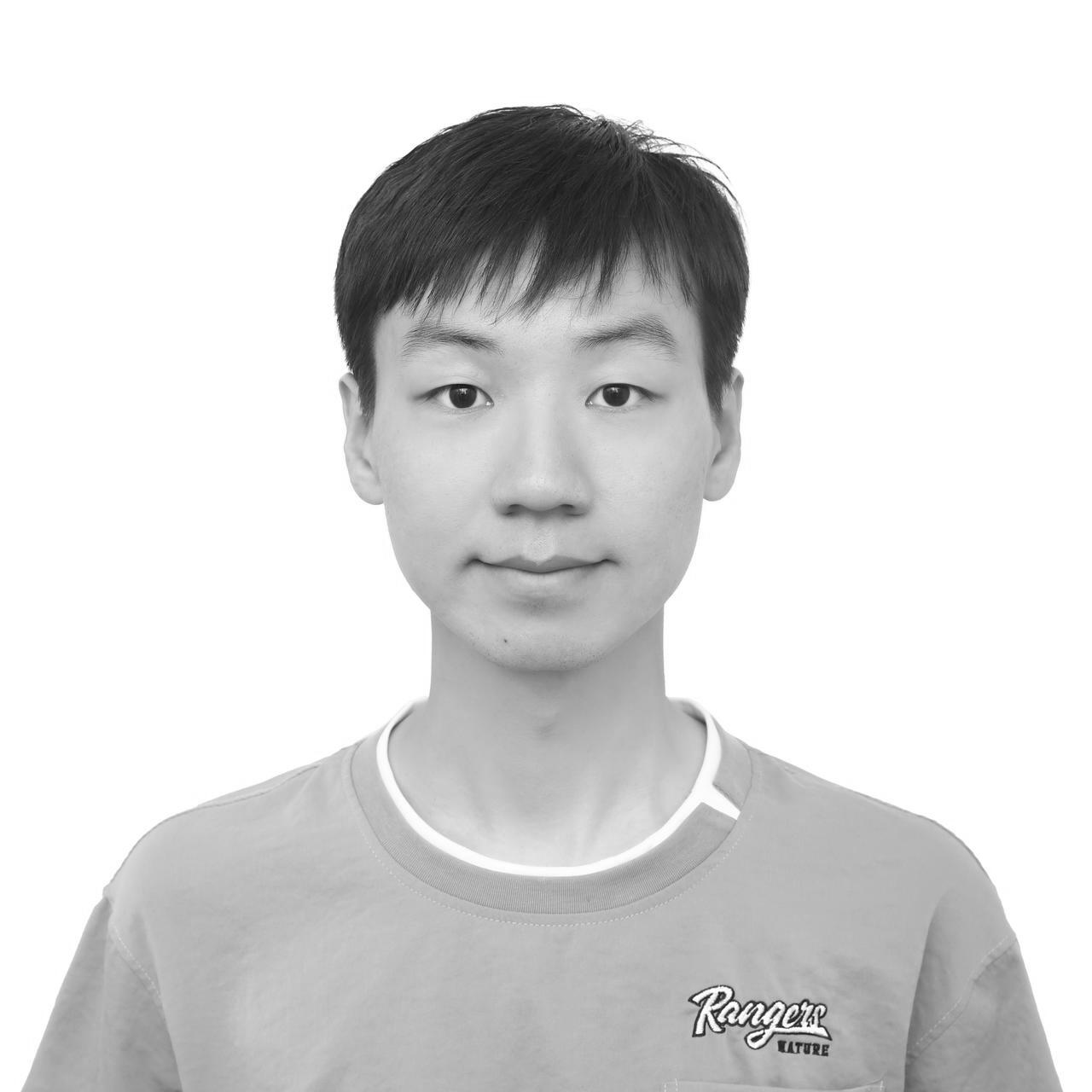}}]
{Xinjie Wang} is a research engineer at Horizon Robotics. 
He received his Master’s degree in Mechanical Automation from Tongji University in 2020. 
His research interests include computer vision and Embodied AI.
\end{IEEEbiography}

\vspace{-.4in}
\begin{IEEEbiography}[{\includegraphics[width=1in]{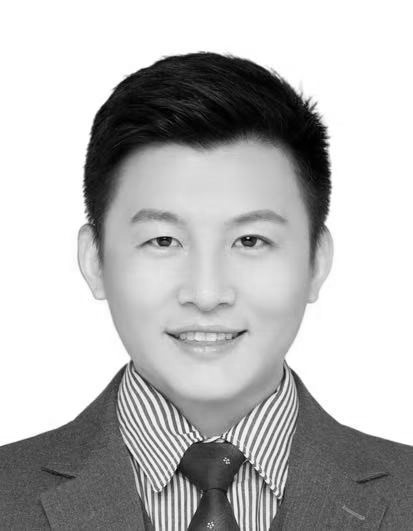}}]
{Wei Sui} is the Algorithm Leader at D-Robotics, responsible for the algorithms 
related to embodied intelligence.
Before that, Dr. Wei Sui was leading the 3D Vision Team at Horizon Robotics. 
His research interests include Structure from Motion (SFM), Simultaneous Localization and Mapping (SLAM), 
Neural Radiance Field (NeRF), 3D perception, etc. Dr. Sui received his B.Eng. and Ph.D. degrees 
from Beihang University and the National Laboratory of Pattern Recognition (NLPR), 
Institute of Automation, Chinese Academy of Sciences (CASIA), Beijing, China, in 2011 and 2016, respectively.
Dr. Wei Sui has published one research monograph and more than ten papers in TIP, TVCG, ICRA, CVPR, etc.  
In addition, he holds over 40 Chinese patents and 5 U.S. patents.
\end{IEEEbiography}

\vspace{-.4in}
\begin{IEEEbiography}[{\includegraphics[width=1in]{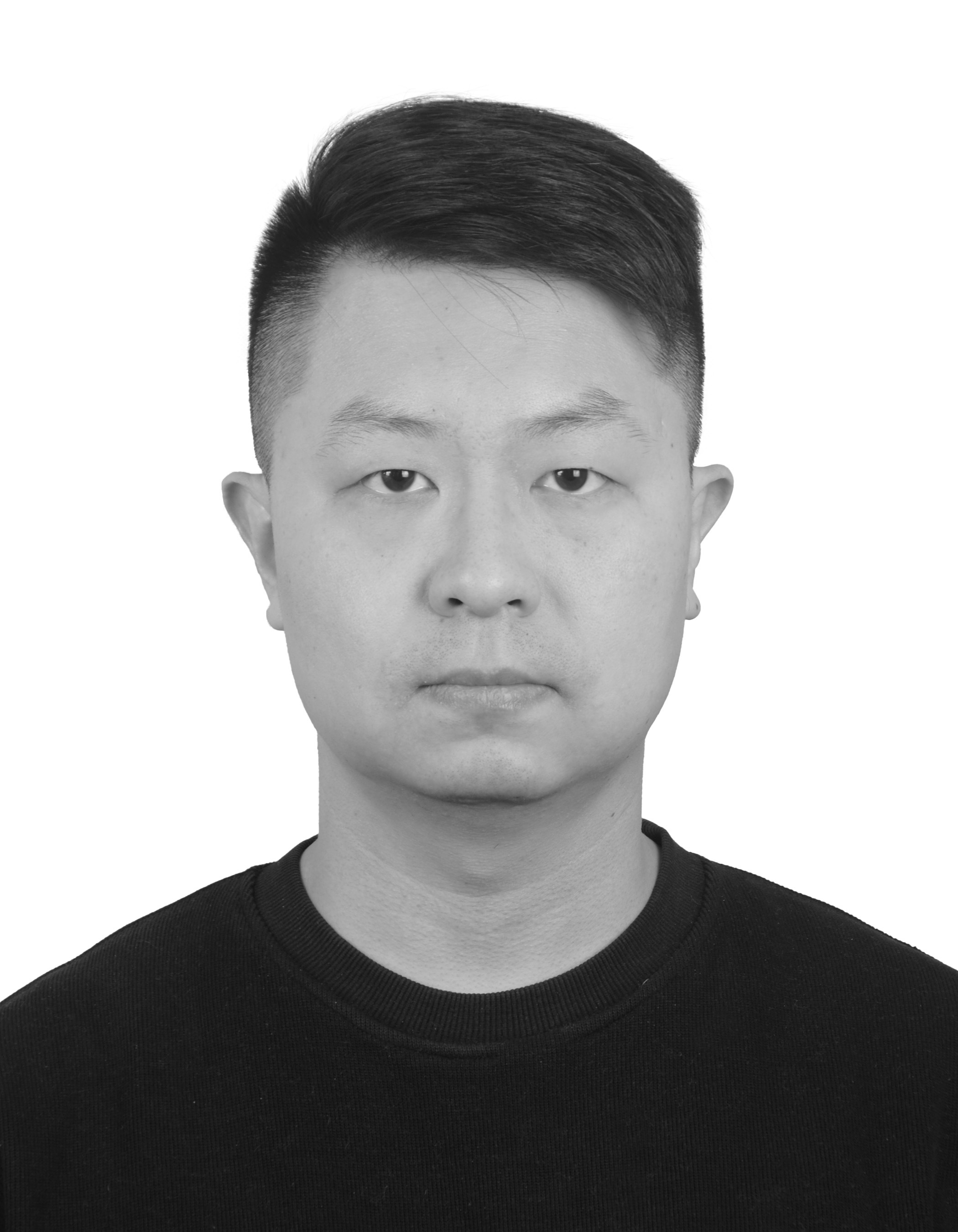}}]
{Zhizhong Su}  is currently the Director of Robot Lab at Horizon Robotics. 
He received his M.S. Degree from Indiana University Bloomington, 
USA and B.E. Degree from Shanghai Jiao Tong University, China. 
His research interests include autonomous driving and robotics learning.
\end{IEEEbiography}

\vspace{-.4in}
\begin{IEEEbiography}[{\includegraphics[width=1in]{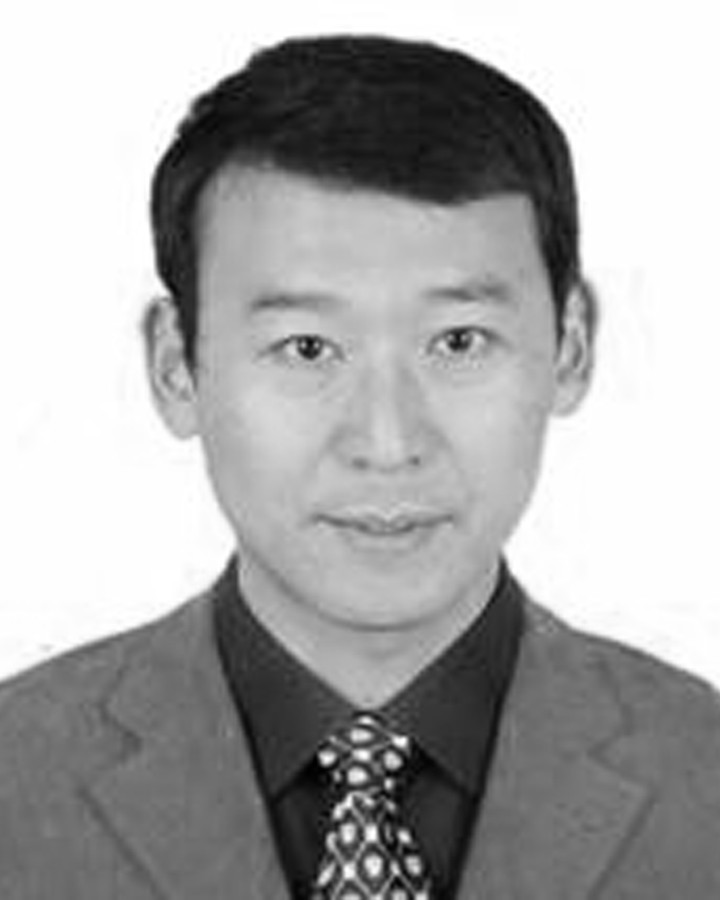}}]
{Jian Yang} received the PhD degree from Nanjing University of Science and Technology (NJUST) in
2002, majoring in pattern recognition and intelligence systems. From 2003 to 2007, he was a
Postdoctoral Fellow at the University of Zaragoza, Hong Kong Polytechnic University and New Jersey
Institute of Technology, respectively. From 2007 to present, he is a professor in the School of Computer
Science and Technology of NJUST. He is the author of more than 300 scientific papers in pattern recognition and computer vision. 
His papers have been cited over 56000 times in the Scholar Google. His research interests include
pattern recognition and computer vision. Currently, he is/was an associate editor of Pattern Recognition, 
Pattern Recognition Letters, IEEE Trans. Neural Networks and Learning Systems, and Neurocomputing. He is a Fellow of IAPR.
\end{IEEEbiography}

\vspace{-.4in}
\begin{IEEEbiography}[{\includegraphics[width=1in]{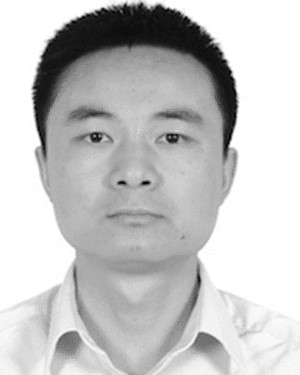}}]
{Jin Xie} received the Ph.D degree in computing from the Department of Computing, The Hong Kong Polytechnic University, Hong Kong, in 2012.
He is currently a Professor with the School of Intelligence Science and Technology, Nanjing University, Nanjing, China. 
He was a Research Scientist with New York University Abu Dhabi from 2013 to 2017. 
Prior to joining Nanjing University in 2023, he was a Professor with the Department of Computer Science and Engineering, 
Nanjing University of Science and Technology, Nanjing, China. His research interests include machine learning, computer vision, computer graphics, and robotics. 
His current research focus is on 3-D computer vision and its applications on autonomous driving and robotic manipulation.
Dr. Xie has authored/co-authored more than 50 papers in well-known journals/conferences such as IEEE TPAMI, IJCV, CVPR, ICCV, ECCV and NeurIPS. 
He has served as a reviewer for IEEE TPAMI, TIP, TNNLS, TMM, CVPR, ICCV and ECCV. 
He was a special issue chair for Asian Conference on Pattern Recognition 2017 and a Guest Editor for Pattern Recognition. 
He was a recipient of the best paper award for Asian Conference on Pattern Recognition 2021.
\end{IEEEbiography}

\vfill

\end{document}